\documentclass{article}

\usepackage{PRIMEarxiv}

\usepackage[utf8]{inputenc} 
\usepackage[T1]{fontenc}    
\usepackage{hyperref}       
\usepackage{url}            
\usepackage{booktabs}       
\usepackage{amsfonts}       
\usepackage{nicefrac}       
\usepackage{microtype}      
\usepackage{lipsum}
\usepackage{fancyhdr}       
\usepackage{graphicx}       
\usepackage[ruled]{algorithm2e}
\usepackage{amssymb}
\usepackage{multirow}
\usepackage{subcaption}
\usepackage{amsmath}
\graphicspath{{media/}}     

\title{Anomaly Multi-classification in Industrial Scenarios: Transferring Few-shot Learning to a New Task}

\author{
  Jie Liu \\
  National Key Laboratory of Human-Machine Hybrid Augmented Intelligence\\
  National Engineering Research Center for Visual Information and Applications\\
  Institute of Artificial Intelligence and Robotics\\
  Xi'an Jiaotong University \\
  \texttt{3121155010@stu.xjtu.edu.cn} \\
   \And
  Yao Wu\\
School of Informatics\\
Xiamen University \\
  \texttt{wuyao@stu.xmu.edu.cn} \\
   \And
  Xiaotong Luo\\
School of Informatics\\
Xiamen University \\
  \texttt{xiaotluo@stu.xmu.edu.cn} \\
  \And 
  Zongze Wu\footnotemark[1]\\
  College of Mechatronics and Control Engineering\\
Shenzhen University\\
\texttt{zzwu@szu.edu.cn}
}

\begin{document}

\maketitle

\begin{abstract}
In industrial scenarios, it is crucial not only to identify anomalous items but also to classify the type of anomaly. However, research on anomaly multi-classification remains largely unexplored. This paper proposes a novel and valuable research task called anomaly multi-classification. Given the challenges in applying few-shot learning to this task, due to limited training data and unique characteristics of anomaly images, we introduce a baseline model that combines RelationNet and PatchCore. We propose a data generation method that creates pseudo classes and a corresponding proxy task, aiming to bridge the gap in transferring few-shot learning to industrial scenarios. Furthermore, we utilize contrastive learning to improve the vanilla baseline, achieving much better performance than directly fine-tune a ResNet. Experiments conducted on MvTec AD and MvTec3D AD demonstrate that our approach shows superior performance in this novel task.

\end{abstract}


\section{Introduction}
In the realm of industrial production, anomaly detection plays a critical role, yet acquiring sufficient defective samples remains a formidable challenge. Various unsupervised anomaly detection methods have emerged. Among the various strategies, PatchCore\cite{roth2022towards} is a notable effective method, leveraging local image information to pinpoint anomalies. Its core concept involves segmenting the image into patches, extracting and modeling features for each patch, and comparing the features of each test sample's patches with those of normal samples to achieve anomaly detection.
\footnotetext[0]{Code will be released in https://github.com/gaoren002/Anomaly-multiclassification}
\footnotetext[1]{Corresponding Author}

Moreover, the field has seen advancements in anomaly detection algorithms tailored for scenarios characterized by a scant number of normal samples, known as few-shot situations. Here, few-shot refers to situations with a small number of normal samples, such as WinCLIP\cite{jeong2023winclip}. It primarily leverages the powerful feature extraction capability of pre-trained CLIP to partition images into multiple windows, using textual prompts to achieve anomaly detection even with limited or zero normal samples. It is important to note, however, that this paper adopts a distinct interpretation of `few-shot' compared to its application in WinCLIP. Few-shot in this paper means few-shot defect samples for multi-classification, rather than few-shot normal samples.

\label{sec:intro}
\begin{figure}[!htbp]
  \centering
  \includegraphics[height=3.5cm]{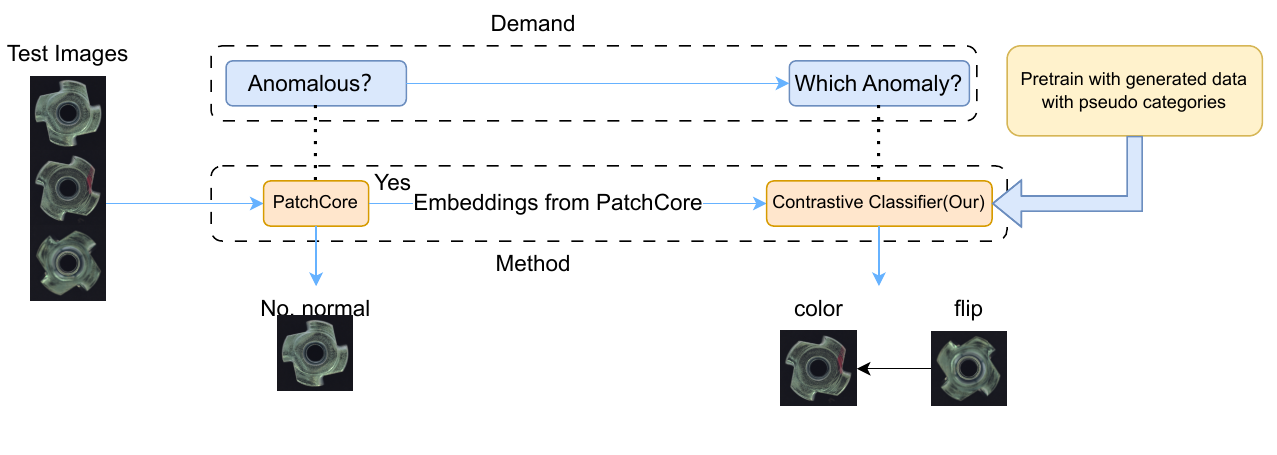}
  \caption{An simple illustration for our method
  }
  \label{fig:structure}
\end{figure}

In practical industrial production, normal samples are usually easy to obtain, while defective samples are hard to acquire. Acknowledging this challenge, we propose a pragmatic task framework that begins with a substantial collection of normal samples, providing only a few samples of different categories of anomalies, the demand is to accurately classify new samples into the specific category of defects. This approach diverges from the conventional anomaly classification paradigm, which is predominantly concerned with the binary determination of defect presence and is essentially a binary classification problem. Our proposed task is a multi-classification problem built upon binary classification assumptions, leveraging the high accuracy of methods like PatchCore to detect all defect samples and then focusing on how to classify defect samples in a few-shot scenario.

Building upon the discussed concepts, the focus shifts to the domain of few-shot learning, a promising avenue for addressing these challenges. Concerning few-shot learning\cite{song2023comprehensive} from the general perspective. In the domain of few-shot image classification, datasets are typically split into training, validation, and test sets. However, the unique aspect of this learning paradigm is that the categories in these three subsets do not overlap. In few-shot learning, each sub-dataset consists of a query set and a support set, both containing data from the same categories, akin to the training and testing sets in traditional tasks.

Most research in this filed is at an early stage, with diverse methods including optimization-based MAML\cite{finn2017model}, graph network-based learning\cite{yang2020dpgn}, and others. This paper focuses on some typical model-based methods such as Siamese Network \cite{koch2015siamese}, Matching Network\cite{vinyals2016matching}, Prototypical Network\cite{snell2017prototypical}, and RelationNet\cite{sung2018learning}. We will build a vanilla baseline based on RelationNet's concept.

Now, the truth is that in industrial scenarios we only have a few defect samples, so that we lack a training dataset that contains different defect classes which do not overlap real defect categories. Thus we can not directly use model-based method in few-shot learning. Faced with this severe scenario, we have to consider few shot learning under the perspective of data generation, by such way, we can easily transfer other few-shot learning method into industrial scenarios, rather than developing new methods especially for industrial scenarios.

Consequently, we introduce a method that seamlessly integrates with anomaly detection processes, and we propose some data generation method to address the lack of defect samples for pretraining in the procedure of few-shot learning. As shown in Figure \ref{fig:structure}. Firstly we detect anomalous image by methods like PatchCore, then we feed the residual representations to a classification model, finally we know which anomaly the test image belong to.

In our endeavor to transpose the insights from few-shot learning research to the sphere of industrial anomaly classification, we have undertaken the following efforts:
\begin{itemize}
\item[$\bullet$] We establish a vanilla baseline based on the RelationNet network for a novel and valuable task called few-shot anomaly multi-classification.
\item[$\bullet$] We propose a data generation method, and utilized them to pre-train the network on proxy tasks, confirming the effectiveness of data generation and proxy tasks in transferring few-shot learning to anomaly classification tasks, thus addressing the issue of lacking training sets while transferring few-shot learning methods into industrial data.
\item[$\bullet$] By using residual representations extracted by PatchCore rather than raw image features, we enhance classification performance and making the vanilla baseline more suitable for industrial anomaly classification tasks.
\item[$\bullet$] We propose some modifications to vanilla baseline inspired by contrastive learning, improving classification performance.
\end{itemize}

\section{Related Work}
\subsection{Anomaly Detection}

Feature-based methods dominate anomaly detection. SPADE\cite{cohen2021subimage} uses the KNN algorithm, leveraging a pre-trained CNN on ImageNet\cite{krizhevsky2012imagenet} for feature extraction, creating a normal instance database. It employs KNN for finding the K closest normal instances and uses feature pyramid matching for pixel-level detection. Similarly, PaDim\cite{defard2021padim} models feature distributions in image patches, while PatchCore\cite{roth2022towards} forms a core set from these patches, clustering normal features with KNN. CPR\cite{li2023target} accelerates matching with histogram similarity, and ReConPatch \cite{hyun2024reconpatch} enhances PatchCore with contrastive learning.

PRN\cite{Zhang_2023_CVPR} detects anomalies by calculating residual features and training a post-processing model, showcasing a hybrid learning approach.

These methods, focusing on binary classification, whose target is to locate the anomaly, differs from our multi-classification approach, whose task is to distinguish the anomaly after we locate the anomaly. A binary classification only need to recognize normal patterns, while multi-classification need to distinguish differences between anomalies. Although similar tasks such as anomaly clustering\cite{sohn2023anomaly} emerge, they require ample defective samples.

\subsection{Few-shot Learning}

Few-shot learning spans knowledge transfer, including parameter sharing and meta-learning, and model-based approaches. Parameter sharing, covering fine-tuning to memory modules, includes Gong et al. 's\cite{gong2022transfer} Faster R-CNN adaptation for X-ray defect detection. Li, Junnan et al.\cite{li2019learning} blend multi-task and few-shot learning, demonstrating the method's versatility. Meta-learning applications, such as Lu et al. 's\cite{lu2020few} autoencoder training, follow a two-phase approach akin to MAML\cite{finn2017model}, showing promise for deployment scenarios. Key models like Siamese\cite{koch2015siamese}, Matching \cite{vinyals2016matching}, Prototypical\cite{snell2017prototypical}, and Relation Networks\cite{sung2018learning} have inspired enhancements for prototype identification, including Hou et al. 's\cite{hou2019cross} cross-attention network. Graph learning, merging with metric learning, introduces novel few-shot learning approaches, exemplified by DPGN\cite{yang2020dpgn}, integrating graph-based learning theory.

However, reliance on extensive pre-training datasets in few-shot learning poses challenges in data-scarce industrial scenarios, so that we can not directly transfer a few-shot learning method into industrial scenarios.
\section{Method}
\subsection{Overview}
\begin{figure}[tb]
  \centering
  \begin{minipage}{0.48\linewidth}
    \centering
    \includegraphics[width=\linewidth]{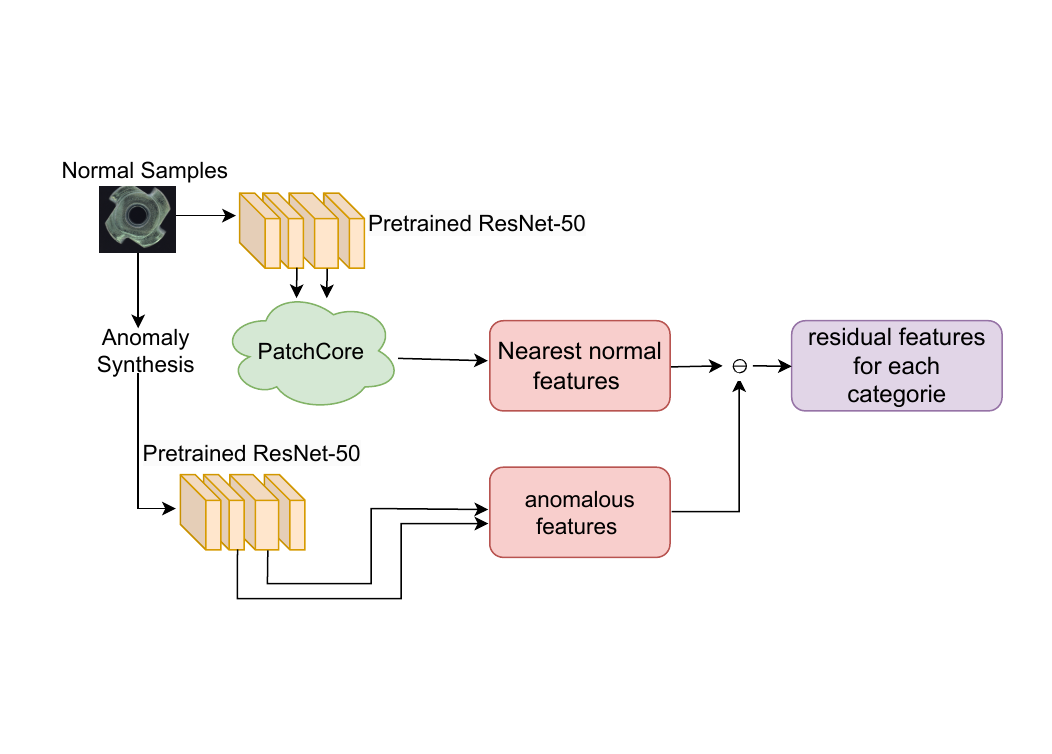}
    \caption{Residual Features Extraction}
    \label{fig:residual}
  \end{minipage}
  \hfill
  \begin{minipage}{0.48\linewidth}
    \centering
    \includegraphics[width=\linewidth]{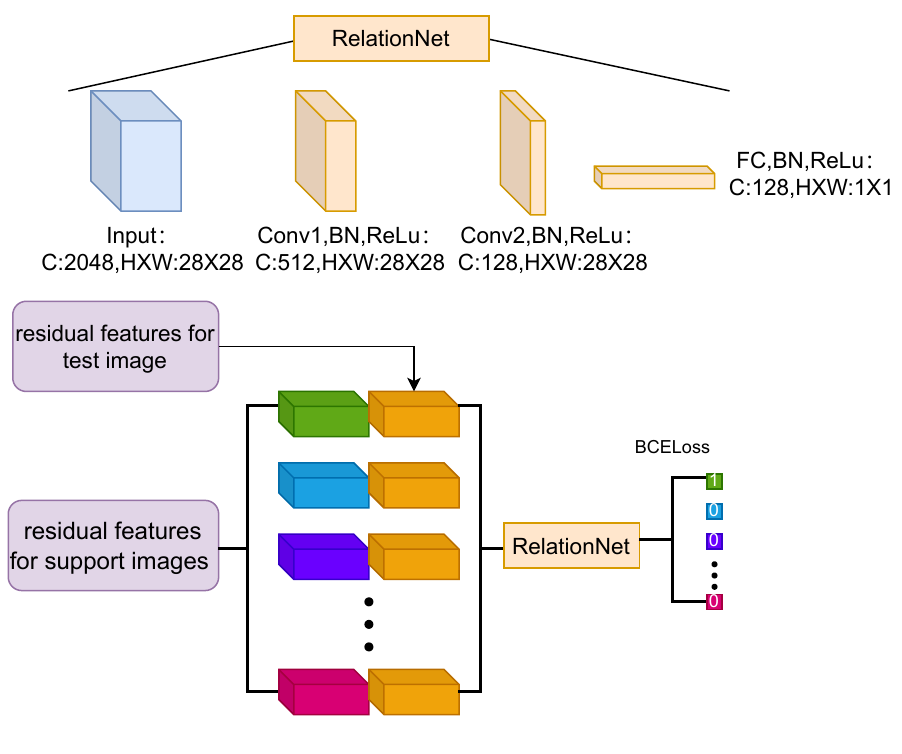}
    \caption{Vanilla Baseline Architecture}
    \label{fig:relationnet}
  \end{minipage}
\end{figure}

In general, our method consists of three modules dedicated to transferring few-shot learning to anomaly multi-classification. First of all, we propose to pretrain with generated data to enhance the performance of our model. The data generation method is shown in Figure \ref{fig:data_generation}. Then we utilize the classic anomaly detection method PatchCore to get the nearest representations of input features, and then we get the residual representations, which can enhance the representation of abnormal areas in anomaly images. As shown in Figure \ref{fig:residual}. Finally we proposed a vanilla baseline for few-shot learning in anomaly multi-classification based on RelationNet and leverage contrastive learning to improve the vanilla baseline. As shown in Figure \ref{fig:relationnet} and Figure \ref{fig:contrastive}.

\subsection{Revisiting the Features Extraction of PatchCore}
\label{subsec:features}
This section follows the setting of PatchCore\cite{roth2022towards}. Assuming $\mathcal{X}$ is the set of normal samples, where the $i$-th sample $X_i \in \mathcal{X}$ is input into a pre-trained model such as ResNet-50\cite{he2016deep}. From the $j$-th layer of the model's output, denoted as $\mathcal{F}_{ij}$, we obtain a feature map. The depth of the feature map tensor $\mathcal{F}_{ij}$ is $c^j$, with height $h^j$ and width $w^j$, i.e., $\mathcal{F}_{ij} \in \mathbb{R}^{c_j \times h_j \times w_j}$. We extract features from the second and third layers, i.e., for each $X_i$, we extract $\mathcal{F}_{i2}\in \mathbb{R}^{c_2 \times h_2 \times w_2}$ and $\mathcal{F}_{i3}\in \mathbb{R}^{c_3 \times h_3 \times w_3}$. Then we patch each position for every sample $X_i$, resulting in feature dimensions of $(w_2h_2, c_2, 3, 3)$ and $(w_3h_3, c_3, 3, 3)$, respectively. 
For the features extracted from layer 3, they are upsampled to $(w_2h_2, c_3, 3, 3)$, then both sets of features undergo adaptive average pooling along the last three dimensions to $c_3$, resulting in features of dimensions $(w_2h_2, c_3)$ for each. Subsequently, these features are concatenated to obtain $(w_2h_2, 2, c_3)$, 
and they are further aggregated(averaged) to obtain $(w_2h_2, c_3)$. For the features of normal samples, all sample features are aggregated into a memory bank $\mathcal{M}$, described by feature dimensions of $(N*w_2h_2, c_3)$, where $N$ denotes the number of normal samples, then memory bank downsampled to $(N_d, c_3)$, where $N_d$ denotes the number of downsampled features. In detail, we define a parameter $p$, where $N_d=N/p$. The downsampled features are sampled from $\mathcal{M}$, and are selected following the coreset subsampling\cite{roth2022towards}in PatchCore. For the features of abnormal or test samples, each image can be described as $A_i (w_2h_2, c_3)\in \mathcal{A}$, with $w_2h_2$ as feature numbers, which $\mathcal{A}$ denotes the set of test samples.
\subsection{Vanilla Baseline}
Inspired by RelationNet, we propose a vanilla baseline as shown in the Figure \ref{fig:relationnet}. In contrast to the original RelationNet, the main difference lies in that we use features extracted in the previous step rather than input images as the input, and then we design a multi-layer convolutional network. Additionally, we modify the original loss function from MSELoss to BCELoss(Equation \ref{eq:my}) and weight the loss to address the imbalance between positive and negative samples. The reason is that it can performs better. Specifically, for the loss of positive sample categories (where features from the support set and query set belong to the same category), we assign a weight of $w=class\_number-1$.
\begin{equation}
\label{eq:my}
L = -\frac{1}{N} \sum_{i=1}^{N} \left[ w \cdot y_i \cdot \log(\hat{y}_i) + (1 - y_i) \cdot \log(1 - \hat{y}_i) \right]
\end{equation}

where $N$ denotes sample number, $y_i$ denotes the label of the sample, $\hat{y}_i$ denotes the prediction of the model
\subsection{Residual Representation}
To further leverage the characteristics of anomaly classification tasks, we propose a enhencement module for the original PatchCore that utilizes residual features. That is, for the features extracted as described in Subsection \ref{subsec:features}, let the features extracted from the input anomalous images be denoted as $A_i(w_2h_2, c_3)$. For any vector whose dimension is $c_3$, we search for the nearest neighbor $M(c_3, )$ in the memory bank $\mathcal{M}$ and then compute the difference as follows:
\begin{equation}
D_i = A_i - M_j \quad where \quad j=\underset{M_j \in \mathcal{M}}{\text{argmin}}\|A_i-M_j\|
\end{equation}

As a result, the input to the vanilla baseline will be residual features, rather than raw features. 
\subsection{Data Generation}
Now, let's review the typical settings of few-shot learning. The categories of items in the training set, as well as those in the testing and validation sets, do not overlap. This condition cannot be met for industrial data because defects of various categories are even scarcer resources and all categories of real defect should be in test set. Therefore, this paper proposes to artificially generate defect images with fabricated categories. A training set is created based on these fabricated defect images, where models can quickly transfer from the defined fabricated categories to real defect categories.

In other words, we define a proxy task by allowing the model to undergo initial classification training on the generated dataset with defined pseudo categories. Then, further fine-tuning training is conducted on few-shot real defect samples. Particularly, in a one-shot scenario where there is insufficient data for fine-tuning, the model trained on the generated dataset can directly classify, yielding satisfactory results in some specific item categories.

\begin{figure}[tb]
  \centering
  \includegraphics[height=5cm]{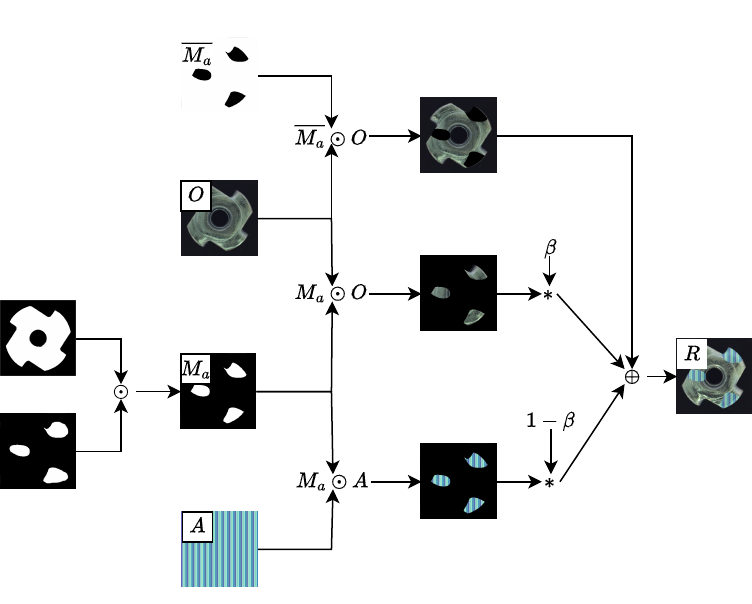}
  \caption{Data Generation Process
  }
  \label{fig:data_generation}
\end{figure}

We define a data generation method, which is inspired by DRAEM\cite{zavrtanik2021draem}, where we also use the DTD\cite{cimpoi2014describing} dataset. This dataset defines 47 different textures. As shown in Figure \ref{fig:data_generation}. Firstly, segmentation methods are employed to obtain foreground-background segmentation masks not present in the original dataset. Backgrounds in industrial images are typically simple, so we employ traditional algorithms such as GrabCut\cite{rother2004grabcut} or flood fill algorithm. Initial seed points are chosen at the perimeter of images, allowing for straightforward extraction of foreground masks. Next, we randomly generate a Poisson noise image and perform threshold segmentation on it to obtain a mask. We then merge this mask with the original image, taking the intersection, to obtain a mask generated solely by Poisson noise on the foreground of the image, denoted as $M_a$. We then merge the texture image with $M_a$, taking the intersection, denoted as $M_a \odot A$. Finally, the resulting image is merged with the original image at a certain ratio $\beta$ to obtain the final fabricated defect image $R$. 
At last we define the category of the fabricated defect image as a category of the texture image. Note that this is not the same as DRAEM, as we remove any augmentation on the texture picture, we should keep the generation method naive to preserve the original texture.

\subsection{Contrastive Classifier}
\begin{figure}[b]
\centering
\includegraphics[height=5cm]{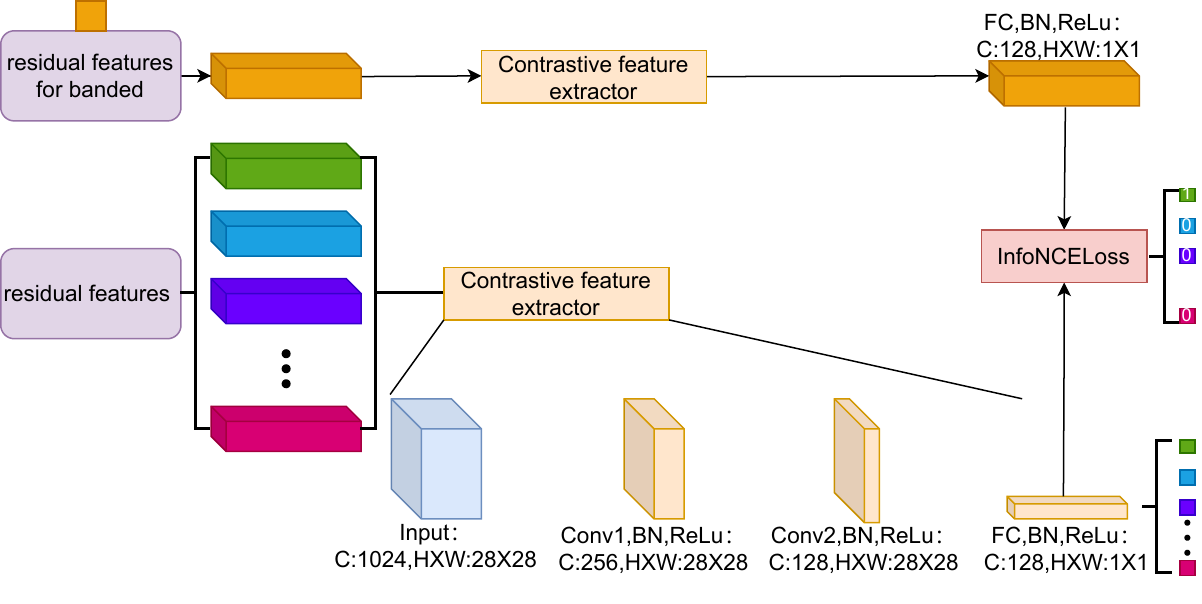}
\caption{Contrastive Learning Architecture}
\label{fig:contrastive}
\end{figure}

Inspired by the success of contrastive learning methods such as SimCLR\cite{chen2020simple} and MOCO\cite{he2020momentum}, we propose further modifications to the vanilla baseline. The modified network architecture is illustrated in Figure \ref{fig:contrastive}.
Specifically, we abandon the original approach used in RelationNet, which concatenates feature vectors and lets the network learn to distinguish the similarity between two sets of features. Instead, we opt for using a multi-layer convolutional network to extract further representations of features. This allows us to obtain representations of the support set and the test set. We then employ InfoNCE\cite{oord2018representation} for contrastive learning. The formula for the InfoNCE loss is as Equation \ref{equa:InfoNCE}:
\begin{equation}
\text{InfoNCE Loss} = -\mathbb{E}_{x_i \in X_s}\left[\log \frac{\exp(f(x_i) \cdot g(x_j))}{\sum_{x_j \in X_t} \exp(f(x_i) \cdot g(x_j))}\right]
\label{equa:InfoNCE}
\end{equation}

Here, $f(x_i)$ and $g(x_j)$ are the residual features of the support set and the test set, respectively. $x_i$ is a sample from the support set $X_s$, representing a category of anomaly, $x_j$ is a sample from the test set $X_t$.

By changing the network architecture and combining it with the InfoNCE loss, we can obtain a better classification network, resulting in better few-shot classification results.

\section{Experiments}
\subsection{Experiments Details}
\subsubsection{Dataset}
We have conducted experiments on the MVTec AD\cite{Bergmann_2019_CVPR} dataset, which comprises fifteen different industrial products, covering various industrial production scenarios. Among them, the category "Toothbrush" is the only one that does not have specific defect categories. The other fourteen categories mostly have three to eight defect categories each, which are the subjects of our study. 

Another dataset MVTev 3D AD\cite{bergmann2021mvtec} is also included in this paper, but we only use the 2D image to classify anomaly.

Other datasets for anomaly detection such as VisA\cite{zou2022spot}, KolektorSDD\cite{tabernik2020segmentation}, BTAD\cite{mishra2021vt}, etc. They don't provide category of anomaly, as we don't include them in this paper.
\subsubsection{Evaluation Metrics}
Since it is a classification problem, the metrics for classification tasks typically include accuracy, precision, recall, etc. Here, for simplicity, we only displayed accuracy. Subsequently, we will calculate the multi-class accuracy for each category of items. The overall classification accuracy is then calculated as the average accuracy of the 14 categories of items.
\subsubsection{Implementation Details}
For the n-shot problem, we select the first n sample from the dataset. Considering the stochastic nature of the downsampling algorithm in PatchCore, different random seeds result in different $\mathcal{M}$ bank, leading to significant fluctuations in classification results. Therefore, for all subsequent experiments involving $\mathcal{M}$, random seeds will be set to 1, 2, 3, 4, and 5, respectively, for five experiments. The average value will be calculated, and the sample standard deviation will be indicated, such as 48.11\% {\tiny $\pm 4.83\%$}.

For the data generation, we only use 1 sample from every category in DTD dataset.

For the vanilla baseline and the Contrastive classifier, different pre-training and training strategies are adopted.

Pre-training involves training with the generated data defined by pseudo-categories. Both pre-training and fine-tuning use a learning rate of 0.0001 and the Adam optimizer.

For both the vanilla baseline and the contrastive classifier, the number of pseudo-categories used for pre-training is 10 (empirical value). Each pre-training iteration randomly selects a specified number of categories from the generated dataset, here is 10. After each iteration, the current training accuracy is recorded, and if the accuracy exceeds the threshold of 0.4 (empirical value), pre-training stops in order to avoid overfitting in generated dataset.

During fine-tuning, for the vanilla baseline, the real data is traversed for training 45 times. For the Contrastive classifier, during fine-tuning, the number of iterations varies depending on the task. For two-shot tasks, the real data is traversed 45 times; for three-shot tasks, it is traversed 25 times, and for four-shot and five-shot tasks, it is traversed 15 times. All of these are empirical value with several simple ablation experiments(not shown in this paper). N-shot means every category of defect has n samples

Both the vanilla baseline and the contrastive classifier require inputs with support sets and query sets simultaneously. For pre-training, with a sufficiently large dataset, samples can be randomly selected from a class of data, and the probability of overlap between support sets and query sets is negligible. For fine-tuning, as the dataset is relatively small, a group of samples needs to be extracted from the input data as the query set, and the remaining data is used as the support set. 
The training details for contrastive classifier is shown in Algorithm \ref{algorithm_contrastive}.

\begin{algorithm}[tb]

\caption{Feature Sorting and Query-Support Pair Generation}
\label{algorithm_contrastive}
\KwData{Input residual features $\mathcal{M}$, shot number $S$, classes number $C$}
\KwResult{For each feature in $\mathcal{M}$, a query sample $\mathcal{Q}$ and its corresponding support set $\mathcal{S}$ are selected}

\BlankLine
\tcp{Sort features by category}
$\mathcal{M}_{sorted} \leftarrow \{\}$\;
\For{$i=1$ \KwTo $C$}{
    \For{$j=1$ \KwTo $S$}{
    Extract $j$th feature of category $i$ from $\mathcal{M}$ and append to $\mathcal{M}_{sorted}$\;
    }
}
$\mathcal{M} \leftarrow \mathcal{M}_{sorted}$\;
\BlankLine
\tcp{Iterate over all features in $\mathcal{M}$}
\For{each feature $\mathcal{F}$ in $\mathcal{M}$}{
    \tcp{Select current feature as the query sample and delete it from the bank}
    $\mathcal{Q} \leftarrow \mathcal{F}$\;
    $\mathcal{M} \leftarrow \mathcal{M} \setminus \mathcal{F}$\;
    \BlankLine
    \tcp{Initialize support set for this query}
    $\mathcal{S} \leftarrow \{\}$\;
    $\mathcal{T} \leftarrow \{\}$\;
    \BlankLine
    \tcp{Select one feature from each category to form the support set}
    \For{$j=1$ \KwTo $S-1$}{
        \For{$i=1$ \KwTo $C$}{
            $\mathcal{F}_{support} \leftarrow$ Select The jth Feature From ith Category($\mathcal{M}$, $i$, $j$)\;
            $\mathcal{S} \leftarrow \mathcal{S} \cup \{\mathcal{F}_{support}\}$\;
        }
        \BlankLine
        \tcp{Now, $\mathcal{Q}$ and $\mathcal{S}$ can be used for training or evaluation}
    }
}
\end{algorithm}

Additionally, the hyper parameters of the PatchCore section remain consistent with the original paper, i.e., the downsampling factor $p=10\%$.

\subsection{Results}
\label{results}
\begin{figure}[t]
  \centering
  \includegraphics[height=6.5cm]{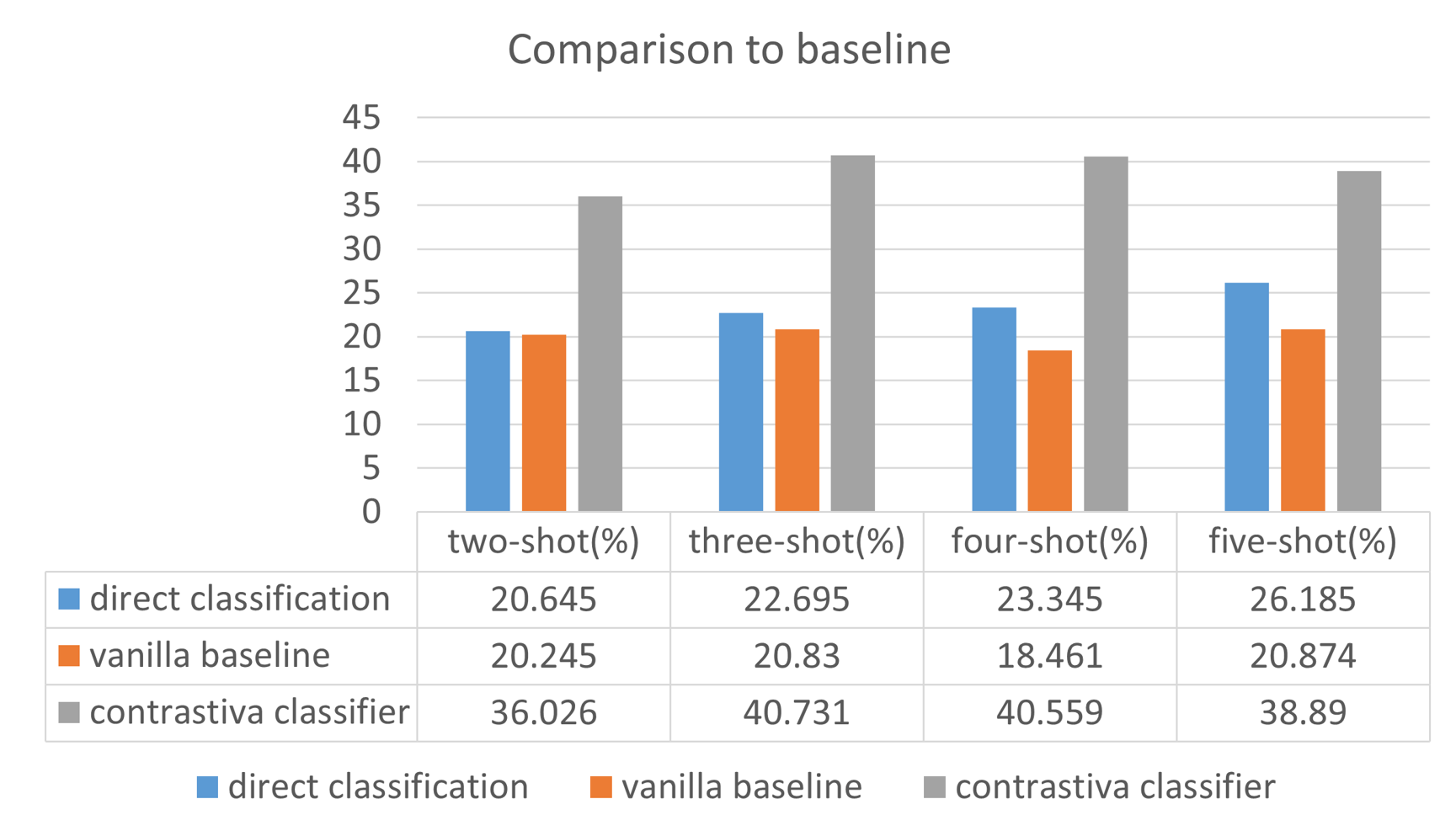}
  \caption{Comparison to baseline
  }
  \label{fig:comparision}
\end{figure}

The comparison results are shown in Figure \ref{fig:comparision}. The classification results of MvTec are shown in Table \ref{tab:contrastive}. Note that we assume all images to be tested are defective. Additionly, it is noted that when the number of shots reaches 4, there is a noticeable anomalous decrease in performance on MvTec AD dataset. Consequently, we also present the results of direct training without pretraining, i.e., without using generated data, in Table \ref{tab:contrastive_without_pretraining}. It can be observed that when the number of shots exceeds 3, the significant gap between our synthesized data and real defects leads to a decrease in performance when using pretraining with synthesized data in four-shot scenarios. However, pretraining with synthesized data still yields better performance in the two-shot, three-shot and five-shot scenarios.

The results of MvTec 3D AD dataset are shown in Table \ref{tab:mvtec3d}.

\begin{table}[tb]
  \centering
  \caption{Contrastive-Classifier Results in MvTec AD Dataset. We especially use bold font if one reaches better performance compared with Table \ref{tab:contrastive_without_pretraining}.}
  \label{tab:contrastive}
  \begin{tabular}{ccccccc}
  \toprule
  category & two-shot(\%) & three-shot(\%) & four-shot(\%) & five-shot(\%) & one-shot(\%)\\
  \midrule
  bottle & {\bf 71.93 {\tiny $\pm 2.48$}} & {\bf 75.92 {\tiny $\pm 1.32$}} & 81.96 {\tiny $\pm 4.25$} & 80.83 {\tiny $\pm 6.82$} & 66.00 {\tiny $\pm 5.08$} \\
  cable & 29.21 {\tiny $\pm 6.13$} & {\bf 35.00 {\tiny $\pm 2.83$}} & 38.67 {\tiny $\pm 4.31$} & {\bf 42.69 {\tiny $\pm 6.29$}} & 13.09 {\tiny $\pm 3.57$}\\
  capsule & {\bf 39.79 {\tiny $\pm 2.63$}} & {\bf 38.72 {\tiny $\pm 4.02$}} & 42.02 {\tiny $\pm 2.33$} & {\bf 47.38 {\tiny $\pm 7.94$}} & 22.12 {\tiny $\pm 5.13$}\\
  carpet & {\bf 27.34 {\tiny $\pm 4.44$}} & {\bf 29.19 {\tiny $\pm 5.46$}} & {\bf 23.77 {\tiny $\pm 0.79$}} & 21.25 {\tiny $\pm 4.22$}  & 23.81 {\tiny $\pm 3.47$}\\
  grid & 21.28 {\tiny $\pm 0.00$} & 19.53 {\tiny $\pm 1.06$} & 15.68 {\tiny $\pm 2.26$} & {\bf 24.38 {\tiny $\pm 6.40$}} & 23.08 {\tiny $\pm 4.51$}\\
  hazelnut & {\bf 42.58 {\tiny $\pm 3.34$}} & {\bf 33.45 {\tiny $\pm 3.13$}} & {\bf 48.52 {\tiny $\pm 6.86$}} & 51.20 {\tiny $\pm 3.03$} & 35.13 {\tiny $\pm 9.67$}\\
  leather & {\bf 29.76 {\tiny $\pm 2.53$}} & {\bf 34.03 {\tiny $\pm 2.32$}} & {\bf 34.44 {\tiny $\pm 2.67$}} & {\bf 38.51 {\tiny $\pm 3.06$}} & 21.84 {\tiny $\pm 3.15$} \\
  metal nut & {\bf 53.17 {\tiny $\pm 1.93$}} & 50.87 {\tiny $\pm 1.35$} &  50.65 {\tiny $\pm 2.06$} & 55.06 {\tiny $\pm 4.27$} & 48.99 {\tiny $\pm 3.43$} \\
  pill & {\bf 34.80 {\tiny $\pm 6.48$}} & 31.00 {\tiny $\pm 4.46$} & 27.61 {\tiny $\pm 1.58$} & {\bf 28.49 {\tiny $\pm 2.94$}} & 22.13 {\tiny $\pm 3.31$} \\
  screw & 20.37 {\tiny $\pm 2.38$} & {\bf 22.31 {\tiny $\pm 1.26$}} & 21.61 {\tiny $\pm 1.53$} & {\bf 22.98 {\tiny $\pm 3.25$}} & 21.41 {\tiny $\pm 3.65$} \\
  tile & {\bf 52.43 {\tiny $\pm 4.42$}} & {\bf 57.10 {\tiny $\pm 6.77$}} & {\bf 59.06 {\tiny $\pm 2.57$}} & 63.05 {\tiny $\pm 1.86$} & 28.61 {\tiny $\pm 5.19$} \\
  transistor & 43.13 {\tiny $\pm 3.42$} & 70.00 {\tiny $\pm 8.22$} & 55.83 {\tiny $\pm 6.32$} & {\bf 66.00{\tiny $\pm 10.84$}} & 31.67 {\tiny $\pm 7.24$} \\
  wood & 24.00 {\tiny $\pm 1.41$} & 29.33 {\tiny $\pm 8.8$} & 40.00{\tiny $\pm 10.90$} & {\bf 42.86 {\tiny $\pm 5.34$}} & 18.55 {\tiny $\pm 5.52$} \\
  zipper & {\bf 30.48 {\tiny $\pm 4.09$}} & 32.04 {\tiny $\pm 5.23$} & {\bf 35.16 {\tiny $\pm 5.10$}} & 30.71 {\tiny $\pm 6.96$} & 14.64 {\tiny $\pm 1.85$} \\
  Average & {\bf 37.16 {\tiny $\pm 0.42$}} & {\bf 39.89 {\tiny $\pm 0.87$}} & 41.07 {\tiny $\pm 1.29$} & {\bf 43.96 {\tiny $\pm 0.90$}} & 27.93 {\tiny $\pm 1.95$} \\
  \bottomrule
\end{tabular}
\end{table}

\begin{table}[tb]
  \centering
  \caption{Contrastive-Classifier Results in MvTec 3D AD Dataset}
  \label{tab:mvtec3d}
  \begin{tabular}{ccccccc}
    \toprule
    category & two-shot(\%) & three-shot(\%) & four-shot(\%) & five-shot(\%) & one-shot(\%)\\
    \midrule
    bagel & 29.50 {\tiny $\pm 1.68$} & 27.37 {\tiny $\pm 1.10$} & 29.45 {\tiny $\pm 1.52$} & 31.47{\tiny $\pm 2.23$} & 24.52 {\tiny $\pm 3.82$} \\
    cable gland & 35.95 {\tiny $\pm 4.79$} & 30.13 {\tiny $\pm 1.20$} & 34.36 {\tiny $\pm 1.61$} & 36.95 {\tiny $\pm 2.00$} & 24.58 {\tiny $\pm 2.78$}\\
    carrot & 27.54 {\tiny $\pm 4.77$} & 31.97 {\tiny $\pm 4.30$} & 34.29 {\tiny $\pm 5.27$} & 35.14 {\tiny $\pm 3.00$} & 29.45 {\tiny $\pm 3.42$}\\
    cookie & 34.53 {\tiny $\pm 3.28$} & 35.39 {\tiny $\pm 4.42$} & 38.29 {\tiny $\pm 3.11$} & 34.70 {\tiny $\pm 1.32$} & 23.63 {\tiny $\pm 2.91$}\\
    dowel & 34.17 {\tiny $\pm 2.26$} & 26.95 {\tiny $\pm 2.70$} &  39.09 {\tiny $\pm 5.83$} &  40.00 {\tiny $\pm 3.43$} & 28.60 {\tiny $\pm 3.51$}\\
    foam & 33.33 {\tiny $\pm 3.93$} &  37.94 {\tiny $\pm 5.24$} &  36.87 {\tiny $\pm 3.77$} & 35.00 {\tiny $\pm 2.36$} & 24.48 {\tiny $\pm 5.23$}\\
    peach & 36.37 {\tiny $\pm 4.17$} & 30.00 {\tiny $\pm 1.75$} &  31.11 {\tiny $\pm 3.33$} & 35.78 {\tiny $\pm 4.27$} & 29.02 {\tiny $\pm 2.03$} \\
    potato & 34.52 {\tiny $\pm 3.95$} & 31.75 {\tiny $\pm 2.88$} &  32.37 {\tiny $\pm 2.56$} &  37.78 {\tiny $\pm 1.81$} & 26.82 {\tiny $\pm 2.74$} \\
    rope & 40.00 {\tiny $\pm 8.79$} & 59.33 {\tiny $\pm 7.13$} & 72.63 {\tiny $\pm 4.57$} &  74.82 {\tiny $\pm 5.94$} & 36.97 {\tiny $\pm 3.49$} \\
    tire & 35.44{\tiny $\pm 10.08$} & 36.80 {\tiny $\pm 3.07$} &  41.41 {\tiny $\pm 3.24$} &  - & 29.64 {\tiny $\pm 6.30$} \\
    Average & 34.14 {\tiny $\pm 0.77$} & 34.76 {\tiny $\pm 1.08$} & 39.00 {\tiny $\pm 1.49$} & 40.18 {\tiny $\pm 1.08$} & 27.77 {\tiny $\pm 1.19$} \\
    \bottomrule
  \end{tabular}
\end{table}

\begin{figure}[tb]
  \centering
  \begin{subfigure}{0.49\linewidth}
    \includegraphics[width=\linewidth]{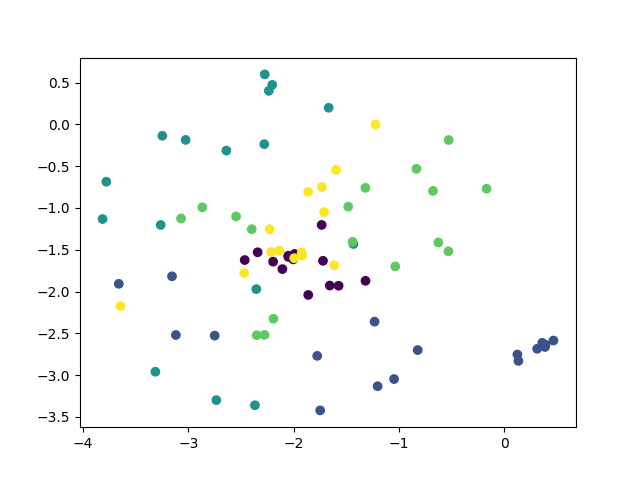}
    \caption{t-sne for tile before contrastive classifier}
    \label{fig:tile-a}
  \end{subfigure}
  \hfill
  \begin{subfigure}{0.49\linewidth}
    \includegraphics[width=\linewidth]{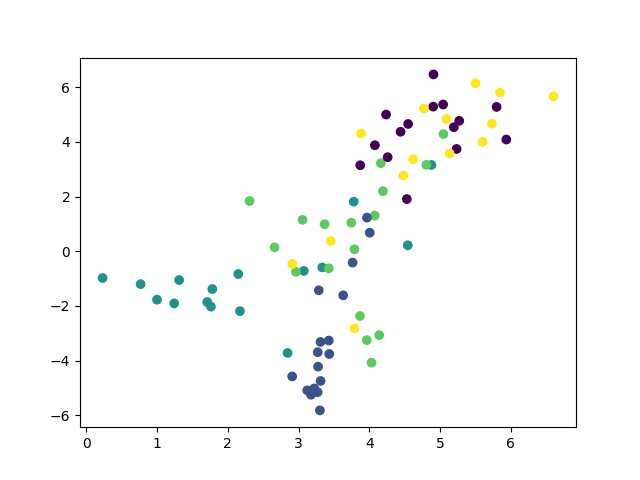}
    \caption{t-sne for tile after contrastive classifier}
    \label{fig:tile-b}
  \end{subfigure}
  \caption{t-sne for tile classification}
  \label{fig:tile}
\end{figure}

\begin{figure}[tb]
  \centering
  \begin{subfigure}{0.49\linewidth}
    \includegraphics[width=\linewidth]{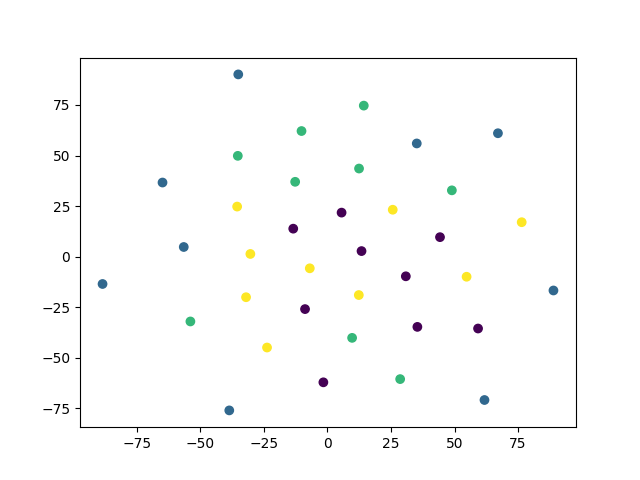}
    \caption{t-sne for transistor before contrastive classifier}
    \label{fig:transistor-a}
  \end{subfigure}
  \hfill
  \begin{subfigure}{0.49\linewidth}
    \includegraphics[width=\linewidth]{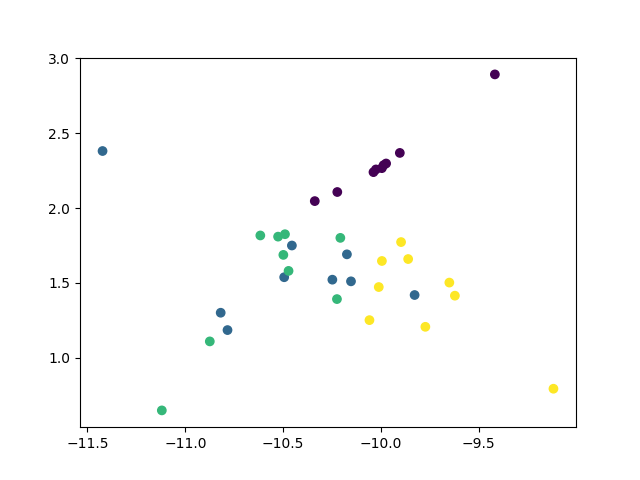}
    \caption{t-sne for transistor after contrastive classifier}
    \label{fig:transistor-b}
  \end{subfigure}
  \caption{t-sne for transistor classification}
  \label{fig:transistor}
\end{figure}

\begin{table}[tb]
  \centering
  \caption{Contrastive-Classifier Results in MVTev AD Dataset Without Pretraining}
  \label{tab:contrastive_without_pretraining}
  \begin{tabular}{cccccc}
    \toprule
    category & two-shot(\%) & three-shot(\%) & four-shot(\%) & five-shot(\%) \\
    \midrule
    bottle & 58.24 {\tiny $\pm 3.59$} & 74.13 {\tiny $\pm 7.28$} & {\bf 85.88 {\tiny $\pm 3.23$}} & {\bf 87.08 {\tiny $\pm 2.28$}}  \\
    cable & {\bf 35.79 {\tiny $\pm 4.78$}} & 33.82 {\tiny $\pm 6.50$} & {\bf 42.67 {\tiny $\pm 7.60$}} & 41.15 {\tiny $\pm 7.27$} \\
    capsule & 38.18 {\tiny $\pm 4.13$} & 36.60 {\tiny $\pm 4.02$} & {\bf 42.47 {\tiny $\pm 3.01$}} &  46.67 {\tiny $\pm 3.98$} \\
    carpet & 21.26 {\tiny $\pm 1.65$} & 21.89 {\tiny $\pm 3.23$} &  23.48 {\tiny $\pm 4.40$} & {\bf 23.44 {\tiny $\pm 2.21$}} \\
    grid & {\bf 23.83 {\tiny $\pm 4.61$}} & {\bf 22.86 {\tiny $\pm 2.71$}} & {\bf 18.38 {\tiny $\pm 4.44$}} & 21.25 {\tiny $\pm 1.40$} \\
    hazelnut & 33.55 {\tiny $\pm 2.65$} & 32.76 {\tiny $\pm 4.04$} & 43.70 {\tiny $\pm 2.81$} & {\bf 51.60 {\tiny $\pm 5.90$}} \\
    leather & 29.27 {\tiny $\pm 2.44$} & 30.91 {\tiny $\pm 4.72$} & 31.94 {\tiny $\pm 4.39$} & 32.84 {\tiny $\pm 5.38$} \\
    metal nut & 49.65 {\tiny $\pm 2.68$} & {\bf 51.61 {\tiny $\pm 2.03$}} & {\bf 51.74 {\tiny $\pm 3.29$}} & {\bf 55.61 {\tiny $\pm 3.44$}} \\
    pill & 29.76 {\tiny $\pm 4.07$} & {\bf 31.33 {\tiny $\pm 2.61$}} & {\bf 31.33 {\tiny $\pm 4.32$}} & {\bf 28.49 {\tiny $\pm 2.53$}} \\
    screw & {\bf 23.12 {\tiny $\pm 1.99$}} & 21.73 {\tiny $\pm 3.94$} & {\bf 22.62 {\tiny $\pm 1.15$}} & {\bf 22.98 {\tiny $\pm3.25$}} \\
    tile & 41.89 {\tiny $\pm 3.82$} & 51.01 {\tiny $\pm 4.27$} & 58.75 {\tiny $\pm 4.89$} & {\bf 64.07 {\tiny $\pm 6.27$}} \\
    transistor & {\bf 49.38 {\tiny $\pm 2.61$}} & {\bf 70.71 {\tiny $\pm 9.91$}} & {\bf 62.50 {\tiny $\pm 6.59$}} & 56.00 {\tiny $\pm 2.24$} \\
    wood & {\bf 34.80 {\tiny $\pm 6.72$}} & {\bf 35.11{\tiny $\pm 10.82$}} & {\bf 45.50 {\tiny $\pm 8.91$}} & {\bf 42.86{\tiny $\pm 11.95$}} \\
    zipper & {\bf 30.48 {\tiny $\pm 4.36$}} & {\bf 34.08 {\tiny $\pm 6.83$}} & 34.72 {\tiny $\pm 4.23$} & {\bf 30.95 {\tiny $\pm 4.98$}} \\
    Average & 35.56 {\tiny $\pm 1.54$} & 39.18 {\tiny $\pm 1.83$} & {\bf 42.55 {\tiny $\pm 2.28$}} &  43.21 {\tiny $\pm 0.91$} \\
    \bottomrule
  \end{tabular}
\end{table}
\begin{table}[tb]
\centering
\caption{Overall Ablation Study}
\label{overall_ablation}
\setlength{\tabcolsep}{9pt} 
\begin{tabular}{cccccc}
    \toprule
    Direct & Baseline & Residual & Pretraining & Contrastive & Results(\%) \\
    \midrule
    \checkmark & & & & & 20.65\\
     & \checkmark& \checkmark & \checkmark & & 27.43\\
     & & & \checkmark & \checkmark& 34.60\\
    & & \checkmark & & \checkmark & 35.56\\
    & & \checkmark& \checkmark& \checkmark& {\bf 37.16}\\
    \bottomrule
\end{tabular}
\end{table}
\begin{table}[htbp]
  \centering
  \begin{minipage}{0.48\textwidth}
   \centering
  \caption{The results of the effectiveness of residual representation}
  \label{tab:residual}
  \begin{tabular}{ccc}
    \toprule
    category & redisual(\%)& without residual(\%)\\
    \midrule
    bottle & {\bf 61.75{\tiny $\pm 11.73$}} &  50.88\\
    cable & {\bf 26.58{\tiny $\pm 10.66$}} & 25.00\\
    capsule & {\bf 35.75 {\tiny $\pm 3.32$}} & 24.24\\
    carpet & {\bf 20.76 {\tiny $\pm 3.18$}} & 20.25\\
    grid & 20.85 {\tiny $\pm 3.49$} & {\bf 34.04}\\
    hazelnut & 49.03{\tiny $\pm 11.94$} & {\bf 50.00}\\
    leather & 25.85 {\tiny $\pm 4.76$} & {\bf 29.27}\\
    matal nut & 49.41 {\tiny $\pm 2.49$} & {\bf 50.59}\\
    pill & {\bf 28.82 {\tiny $\pm 7.06$}} & 18.90\\
    screw & {\bf 22.57 {\tiny $\pm 1.54$}} & 22.02\\
    tile & {\bf 48.11 {\tiny $\pm 4.83$}} & 35.14\\
    transistor & 50.00{\tiny $\pm 11.48$} & {\bf 56.25}\\
    wood & 36.00 {\tiny $\pm 7.35$} & {\bf 44.00}\\
    zipper & {\bf 26.48 {\tiny $\pm 3.83$}} & 23.81\\
    Average & {\bf 36.03} & 34.60\\
    \bottomrule
  \end{tabular}
  \end{minipage}
  \hfill
  \begin{minipage}{0.49\textwidth}
 \centering
  \caption{The effectiveness of contrastive learning}
  \label{tab:effectiveness_contrastive}
  \begin{tabular}{ccc}
    \toprule
    category & contrastive(\%) & vanilla baseline(\%)\\
    \midrule
    bottle &  58.95 {\tiny $\pm 8.82$} & {\bf 61.41 {\tiny $\pm 4.96$}}\\
    cable & {\bf 26.58{\tiny $\pm 10.66$}} & 15.53 {\tiny $\pm 3.27$}\\
    capsule & {\bf 35.75 {\tiny $\pm 3.32$}} & 22.02 {\tiny $\pm 2.41$}\\
    carpet & 20.76 {\tiny $\pm 3.18$} & {\bf 24.05 {\tiny $\pm 5.30$}}\\
    grid & {\bf 20.85 {\tiny $\pm 3.49$}} & 18.72 {\tiny $\pm 3.16$}\\
    hazelnut & {\bf 49.03{\tiny $\pm 11.94$}} & 33.23 {\tiny $\pm 3.34$}\\
    leather & {\bf 25.85 {\tiny $\pm 4.76$}} &  21.22 {\tiny $\pm 6.77$}\\
    metal nut & {\bf 49.41 {\tiny $\pm 2.49$}} & 47.76 {\tiny $\pm 4.52$} \\
    pill & {\bf 28.82 {\tiny $\pm 7.06$}} & 20.63 {\tiny $\pm 4.18$}\\
    screw &  {\bf 22.57 {\tiny $\pm 1.54$}} & 19.27 {\tiny $\pm 2.34$}\\
    tile &  {\bf 48.11 {\tiny $\pm 4.83$}} &  35.41 {\tiny $\pm 6.37$}\\
    transistor & {\bf 50.00{\tiny $\pm 11.48$}} & 32.51 {\tiny $\pm 8.15$}\\
    wood & {\bf 36.00 {\tiny $\pm 7.35$}} & 14.40 {\tiny $\pm 4.98$}\\
    zipper & {\bf 26.48 {\tiny $\pm 3.83$}} & 15.62 {\tiny $\pm 1.86$}\\
    Average & {\bf 36.03} & 27.43\\
    \bottomrule
  \end{tabular}
  \end{minipage}
\end{table}
Additionally, to demonstrate that the contrastive classifier we use indeed explicitly represents the features in an appropriate manner, we present the t-SNE plots of two categories in two-shot scenarios before and after contrastive classifier, as shown in Figure \ref{fig:tile} and Figure \ref{fig:transistor}.
\subsection{Ablation Study}
The overall ablation study is shown as Table \ref{overall_ablation}, we only show the two-shot scenario's result. The direct classification means we fine-tune the last fully connected layer of ResNet-50. As shown, the contrastive classifier with residual representation and pretraining achieves the best results in contrast to other methods.

\subsubsection{Data Generation and Pretraining}
As shown in Table \ref{tab:contrastive} and Table \ref{tab:contrastive_without_pretraining}. The results are as we have discussed in Subsection \ref{results}, in two-shot, three-shot and five-shot scenarios, pretraining with generated data shows better performance. 
\subsubsection{Residual Representations}
In two-shot scenarios, we use contrasitive classifier with pretraning to compare the results between utilizing residual representations and not utilizing residual representations. The result is shown in Table \ref{tab:residual}. As we can see, we achieve better results when we use 
residual representations.
\subsubsection{The Effectiveness of Contrasitive Classifier}
In two-shot scenarios, we compare contrasitive classifier with pretraning to vanilla baseline with pretraining. And residual representations are adopted. The result is shown in Table \ref{tab:effectiveness_contrastive}. The contrastive classifier achieve better results in 11 items, and increase about 8\% in average accuracy, which shows the effectiveness of contrastive learning for classifier.

\section{Conclusion}

In this paper, we explore the necessity of anomaly classification following anomaly detection, proposing a novel and valuable research task. We introduce a vanilla baseline for this task. Subsequently, in order to transfer few-shot learning to industrial applications, we propose a data generation method to enhance the model's classification capability, achieving better results in two-shot and three-shot and five-shot scenarios. Furthermore, our proposed residual representation and contrastive learning based improvements to the vanilla baseline further enhance the model's classification ability, improving classification accuracy.

Besides, our method can be easily embedded into any anomaly detection algorithm similar to PatchCore, e.g., Padim, SPADE, CPR.
\section{Limitations}
The data generation method proposed in this paper fails when there are four samples, indicating significant room for improvement in data generation methods for industrial data. We should anticipate more research in this area in the future.

Moreover, overall, the classification accuracy for defects in certain categories remains at the level of random guessing, indicating the difficulty of this task. Further research, such as metric-based or optimization-based methods, can be used to explore few-shot learning on industrial data.
\bibliographystyle{unsrt}
\bibliography{references}

\begin{thebibliography}{10}

\bibitem{roth2022towards}
Karsten Roth, Latha Pemula, Joaquin Zepeda, Bernhard Sch{\"o}lkopf, Thomas
  Brox, and Peter Gehler.
\newblock Towards total recall in industrial anomaly detection.
\newblock In {\em CVPR}, pages 14318--14328, 2022.

\bibitem{jeong2023winclip}
Jongheon Jeong, Yang Zou, Taewan Kim, Dongqing Zhang, Avinash Ravichandran, and
  Onkar Dabeer.
\newblock Winclip: Zero-/few-shot anomaly classification and segmentation.
\newblock In {\em CVPR}, pages 19606--19616, 2023.

\bibitem{song2023comprehensive}
Yisheng Song, Ting Wang, Puyu Cai, Subrota~K Mondal, and Jyoti~Prakash Sahoo.
\newblock A comprehensive survey of few-shot learning: Evolution, applications,
  challenges, and opportunities.
\newblock {\em ACM Computing Surveys}, 2023.

\bibitem{finn2017model}
Chelsea Finn, Pieter Abbeel, and Sergey Levine.
\newblock Model-agnostic meta-learning for fast adaptation of deep networks.
\newblock In {\em ICML}, pages 1126--1135. PMLR, 2017.

\bibitem{yang2020dpgn}
Ling Yang, Liangliang Li, Zilun Zhang, Xinyu Zhou, Erjin Zhou, and Yu~Liu.
\newblock Dpgn: Distribution propagation graph network for few-shot learning.
\newblock In {\em CVPR}, pages 13390--13399, 2020.

\bibitem{koch2015siamese}
Gregory Koch, Richard Zemel, Ruslan Salakhutdinov, et~al.
\newblock Siamese neural networks for one-shot image recognition.
\newblock In {\em ICML deep learning workshop}, volume~2. Lille, 2015.

\bibitem{vinyals2016matching}
Oriol Vinyals, Charles Blundell, Timothy Lillicrap, Daan Wierstra, et~al.
\newblock Matching networks for one shot learning.
\newblock {\em NeurIPS}, 29, 2016.

\bibitem{snell2017prototypical}
Jake Snell, Kevin Swersky, and Richard Zemel.
\newblock Prototypical networks for few-shot learning.
\newblock {\em NeurIPS}, 30, 2017.

\bibitem{sung2018learning}
Flood Sung, Yongxin Yang, Li~Zhang, Tao Xiang, Philip~HS Torr, and Timothy~M
  Hospedales.
\newblock Learning to compare: Relation network for few-shot learning.
\newblock In {\em CVPR}, pages 1199--1208, 2018.

\bibitem{cohen2021subimage}
Niv Cohen and Yedid Hoshen.
\newblock Sub-image anomaly detection with deep pyramid correspondences, 2021.

\bibitem{krizhevsky2012imagenet}
Alex Krizhevsky, Ilya Sutskever, and Geoffrey~E Hinton.
\newblock Imagenet classification with deep convolutional neural networks.
\newblock {\em NeurIPS}, 25, 2012.

\bibitem{defard2021padim}
Thomas Defard, Aleksandr Setkov, Angelique Loesch, and Romaric Audigier.
\newblock Padim: a patch distribution modeling framework for anomaly detection
  and localization.
\newblock In {\em ICPR}, pages 475--489. Springer, 2021.

\bibitem{li2023target}
Hanxi Li, Jianfei Hu, Bo~Li, Hao Chen, Yongbin Zheng, and Chunhua Shen.
\newblock Target before shooting: Accurate anomaly detection and localization
  under one millisecond via cascade patch retrieval.
\newblock {\em arXiv preprint arXiv:2308.06748}, 2023.

\bibitem{hyun2024reconpatch}
Jeeho Hyun, Sangyun Kim, Giyoung Jeon, Seung~Hwan Kim, Kyunghoon Bae, and
  Byung~Jun Kang.
\newblock Reconpatch: Contrastive patch representation learning for industrial
  anomaly detection.
\newblock In {\em CVPR}, pages 2052--2061, 2024.

\bibitem{Zhang_2023_CVPR}
Hui Zhang, Zuxuan Wu, Zheng Wang, Zhineng Chen, and Yu-Gang Jiang.
\newblock Prototypical residual networks for anomaly detection and
  localization.
\newblock In {\em CVPR}, pages 16281--16291, June 2023.

\bibitem{sohn2023anomaly}
Kihyuk Sohn, Jinsung Yoon, Chun-Liang Li, Chen-Yu Lee, and Tomas Pfister.
\newblock Anomaly clustering: Grouping images into coherent clusters of anomaly
  types.
\newblock In {\em WACV}, pages 5479--5490, 2023.

\bibitem{gong2022transfer}
Yanfeng Gong, Jun Luo, Hongliang Shao, and Zhixue Li.
\newblock A transfer learning object detection model for defects detection in
  x-ray images of spacecraft composite structures.
\newblock {\em Composite Structures}, 284:115136, 2022.

\bibitem{li2019learning}
Junnan Li, Yongkang Wong, Qi~Zhao, and Mohan~S Kankanhalli.
\newblock Learning to learn from noisy labeled data.
\newblock In {\em CVPR}, pages 5051--5059, 2019.

\bibitem{lu2020few}
Yiwei Lu, Frank Yu, Mahesh Kumar~Krishna Reddy, and Yang Wang.
\newblock Few-shot scene-adaptive anomaly detection.
\newblock In {\em ECCV}, pages 125--141. Springer, 2020.

\bibitem{hou2019cross}
Ruibing Hou, Hong Chang, Bingpeng Ma, Shiguang Shan, and Xilin Chen.
\newblock Cross attention network for few-shot classification.
\newblock {\em NeurIPS}, 32, 2019.

\bibitem{he2016deep}
Kaiming He, Xiangyu Zhang, Shaoqing Ren, and Jian Sun.
\newblock Deep residual learning for image recognition.
\newblock In {\em CVPR}, pages 770--778, 2016.

\bibitem{zavrtanik2021draem}
Vitjan Zavrtanik, Matej Kristan, and Danijel Sko{\v{c}}aj.
\newblock Draem-a discriminatively trained reconstruction embedding for surface
  anomaly detection.
\newblock In {\em ICCV}, pages 8330--8339, 2021.

\bibitem{cimpoi2014describing}
Mircea Cimpoi, Subhransu Maji, Iasonas Kokkinos, Sammy Mohamed, and Andrea
  Vedaldi.
\newblock Describing textures in the wild.
\newblock In {\em CVPR}, pages 3606--3613, 2014.

\bibitem{rother2004grabcut}
Carsten Rother, Vladimir Kolmogorov, and Andrew Blake.
\newblock " grabcut" interactive foreground extraction using iterated graph
  cuts.
\newblock {\em TOG}, 23(3):309--314, 2004.

\bibitem{chen2020simple}
Ting Chen, Simon Kornblith, Mohammad Norouzi, and Geoffrey Hinton.
\newblock A simple framework for contrastive learning of visual
  representations.
\newblock In {\em ICML}, pages 1597--1607. PMLR, 2020.

\bibitem{he2020momentum}
Kaiming He, Haoqi Fan, Yuxin Wu, Saining Xie, and Ross Girshick.
\newblock Momentum contrast for unsupervised visual representation learning.
\newblock In {\em CVPR}, pages 9729--9738, 2020.

\bibitem{oord2018representation}
Aaron van~den Oord, Yazhe Li, and Oriol Vinyals.
\newblock Representation learning with contrastive predictive coding.
\newblock {\em arXiv preprint arXiv:1807.03748}, 2018.

\bibitem{Bergmann_2019_CVPR}
Paul Bergmann, Michael Fauser, David Sattlegger, and Carsten Steger.
\newblock Mvtec ad -- a comprehensive real-world dataset for unsupervised
  anomaly detection.
\newblock In {\em CVPR}, June 2019.

\bibitem{bergmann2021mvtec}
Paul Bergmann, Xin Jin, David Sattlegger, and Carsten Steger.
\newblock The mvtec 3d-ad dataset for unsupervised 3d anomaly detection and
  localization.
\newblock {\em arXiv preprint arXiv:2112.09045}, 2021.

\bibitem{zou2022spot}
Yang Zou, Jongheon Jeong, Latha Pemula, Dongqing Zhang, and Onkar Dabeer.
\newblock Spot-the-difference self-supervised pre-training for anomaly
  detection and segmentation.
\newblock In {\em European Conference on Computer Vision}, pages 392--408.
  Springer, 2022.

\bibitem{tabernik2020segmentation}
Domen Tabernik, Samo {\v{S}}ela, Jure Skvar{\v{c}}, and Danijel Sko{\v{c}}aj.
\newblock Segmentation-based deep-learning approach for surface-defect
  detection.
\newblock {\em Journal of Intelligent Manufacturing}, 31(3):759--776, 2020.

\bibitem{mishra2021vt}
Pankaj Mishra, Riccardo Verk, Daniele Fornasier, Claudio Piciarelli, and
  Gian~Luca Foresti.
\newblock Vt-adl: A vision transformer network for image anomaly detection and
  localization.
\newblock In {\em 2021 IEEE 30th International Symposium on Industrial
  Electronics (ISIE)}, pages 01--06. IEEE, 2021.

\end{thebibliography}

\appendix

\newpage
\begin{center}
    \Large \textbf{Appendix}
\end{center}

\section{Vanilla baseline Training Algorithm}
Vanilla baseline can be difficult to train, we can train it similarly to Algorithm \ref{algorithm_contrastive}, but we found that it does not learn anything. Thus we apply a simpler training strategy described in Algorithm \ref{algorithm_vanilla}.

\section{The Complete Results of Vanilla Baseline}
As shown in Table \ref{tab:vanilla}. We can find that the vanilla baseline is hard to train and doesn't perform better as the number of samples increases. The results of vanilla baseline without pretraining are shown in Table \ref{tab:vanilla_without_pretraining}. Pretraining makes it performs better in two-shot scenario.

\section{Another data generation method}
The second method of image generation is largely similar to the first method, with the main difference being that the generated masks are specific-sized n-sided polygons, where $n\in {3, 4, 5, 6}$. For texture images belonging to the same category from DTD, if they correspond to different-shaped masks, they will be labeled as different categories accordingly. Consequently, a maximum of $47*4$ 
defect images can be generated in this way. The initial purpose of this approach is to encourage the model to learn to distinguish between pseudo-defect texture information while also paying attention to the shape and edges of the defects. The results are shown in Table \ref{tab:new
  _data_generation}. But it doesn't show better performance even worse than direct training due to overfitting in generated data.

\section{Other Results of Direct Classification}
we will show the complete results here. As shown in Table \ref{tab:direct}

\section{The Complete Results of Contrastive Classifier without Residual Representation}
Results are shown in Table \ref{tab:without_residual}

\section{MvTec 3D AD Results including good samples}
In MvTec, PatchCore can achieve very high performance, while in MvTec 3D when we only use the RGB image it sometimes fails. Therefore in the whole workflow, there can be some good samples mixed with anomalous samples to be classified. We will show the results considerring normal samples in Table \ref{tab:mvtec3d_good}.

\section{Combine MAML with our model}
We attempted some optimization-based few-shot learning methods, which are generally model-agnostic, such as MAML. The operational method is similar to MAML, but there is a slight difference due to our model inherently incorporating both the support set and the query set. This differs from MAML, which performs a first gradient descent on the support set before conducting a second gradient descent on the query set. Therefore, in our generated dataset, we selected a batch of categories as the support set for the first gradient descent and another batch of categories as the query set for the second gradient descent. This approach is consistent with the original concept of MAML and has been found to achieve better results in the two-shot scenario. The experimental data are shown in Table \ref{tab:maml}.

\section{Complete t-SNE results}
\begin{figure}[tb]
  \centering
  \begin{subfigure}{0.49\linewidth}
    \includegraphics[width=\linewidth]{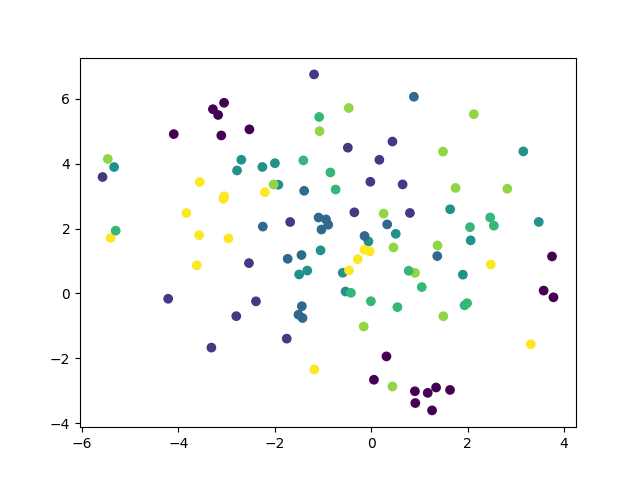}
    \caption{t-sne for zipper before contrastive classifier}
    \label{fig:zipper-a}
  \end{subfigure}
  \hfill
  \begin{subfigure}{0.49\linewidth}
    \includegraphics[width=\linewidth]{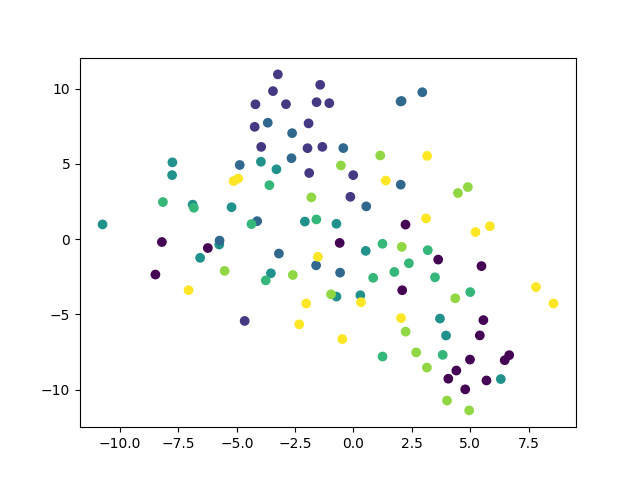}
    \caption{t-sne for zipper after contrastive classifier}
    \label{fig:zipper-b}
  \end{subfigure}
  \caption{t-sne for zipper classification}
  \label{fig:zipper}
\end{figure}

\begin{figure}[tb]
  \centering
  \begin{subfigure}{0.49\linewidth}
    \includegraphics[width=\linewidth]{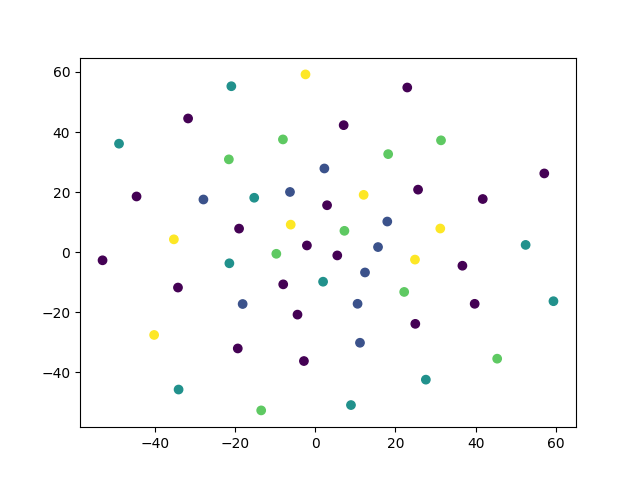}
    \caption{t-sne for wood before contrastive classifier}
    \label{fig:wood-a}
  \end{subfigure}
  \hfill
  \begin{subfigure}{0.49\linewidth}
    \includegraphics[width=\linewidth]{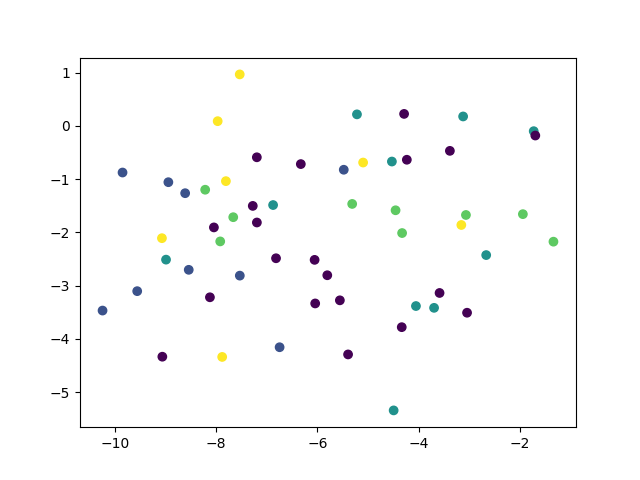}
    \caption{t-sne for wood after contrastive classifier}
    \label{fig:wood-b}
  \end{subfigure}
  \caption{t-sne for wood classification}
  \label{fig:wood}
\end{figure}

\begin{figure}[tb]
  \centering
  \begin{subfigure}{0.49\linewidth}
    \includegraphics[width=\linewidth]{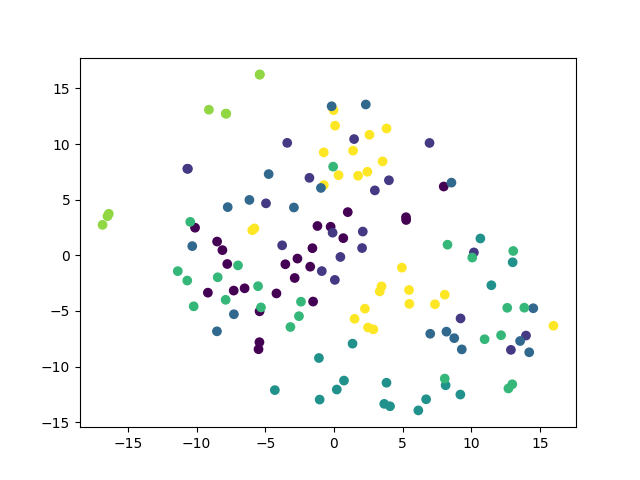}
    \caption{t-sne for pill before contrastive classifier}
    \label{fig:pill-a}
  \end{subfigure}
  \hfill
  \begin{subfigure}{0.49\linewidth}
    \includegraphics[width=\linewidth]{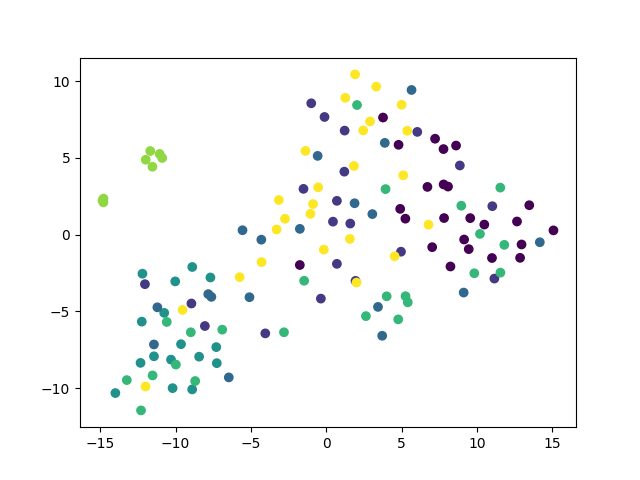}
    \caption{t-sne for pill after contrastive classifier}
    \label{fig:pill-b}
  \end{subfigure}
  \caption{t-sne for pill classification}
  \label{fig:pill}
\end{figure}

\begin{figure}[tb]
  \centering
  \begin{subfigure}{0.49\linewidth}
    \includegraphics[width=\linewidth]{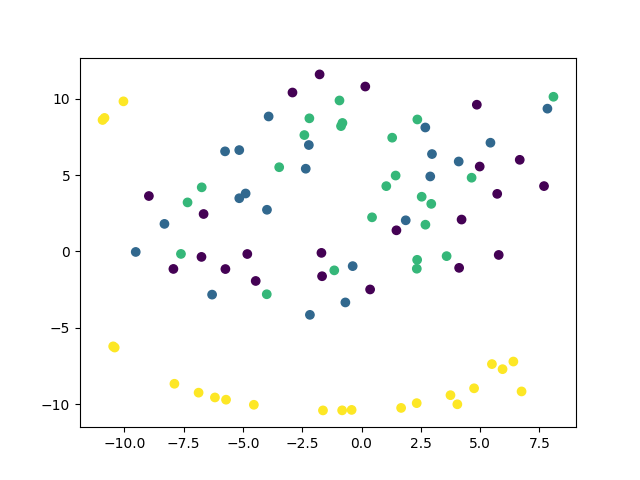}
    \caption{t-sne for metal nut before contrastive classifier}
    \label{fig:metal_nut-a}
  \end{subfigure}
  \hfill
  \begin{subfigure}{0.49\linewidth}
    \includegraphics[width=\linewidth]{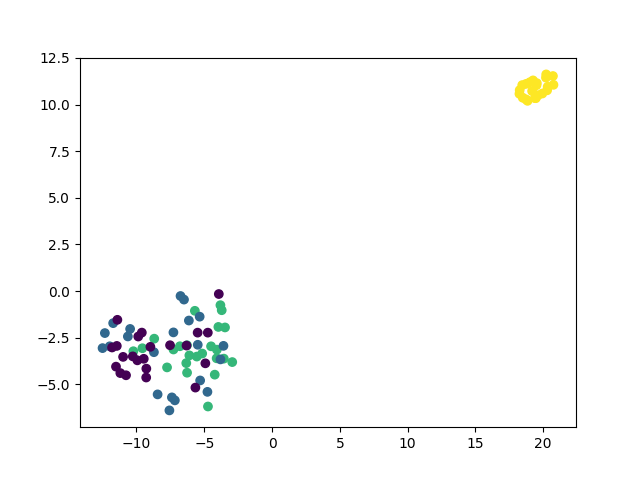}
    \caption{t-sne for metal nut after contrastive classifier}
    \label{fig:metal_nut-b}
  \end{subfigure}
  \caption{t-sne for metal nut classification}
  \label{fig:metal_nut}
\end{figure}

\begin{figure}[tb]
  \centering
  \begin{subfigure}{0.49\linewidth}
    \includegraphics[width=\linewidth]{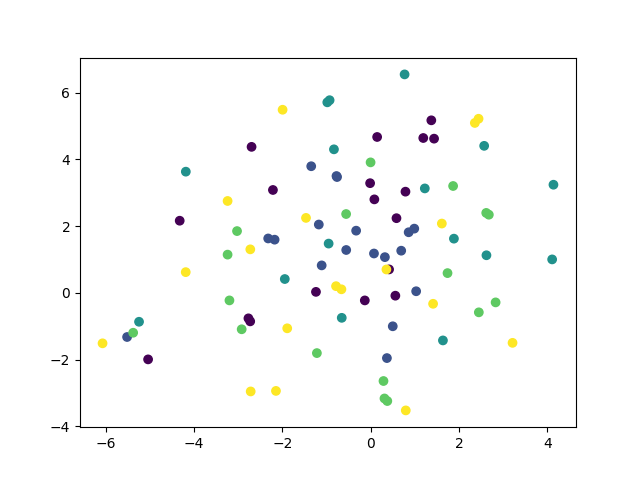}
    \caption{t-sne for leather before contrastive classifier}
    \label{fig:leather-a}
  \end{subfigure}
  \hfill
  \begin{subfigure}{0.49\linewidth}
    \includegraphics[width=\linewidth]{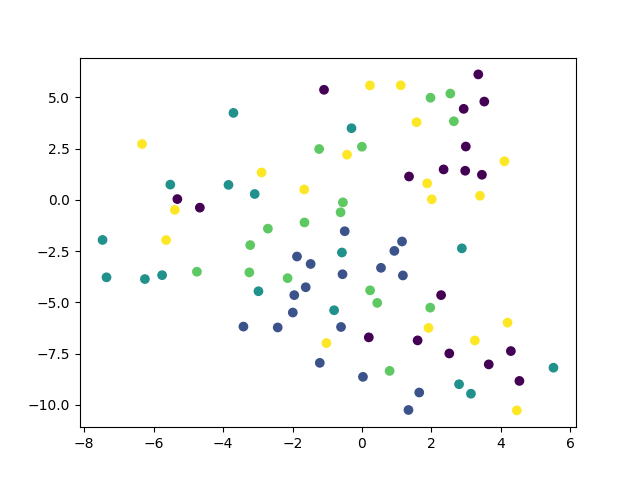}
    \caption{t-sne for leather after contrastive classifier}
    \label{fig:leather-b}
  \end{subfigure}
  \caption{t-sne for leather classification}
  \label{fig:leather}
\end{figure}

\begin{figure}[tb]
  \centering
  \begin{subfigure}{0.49\linewidth}
    \includegraphics[width=\linewidth]{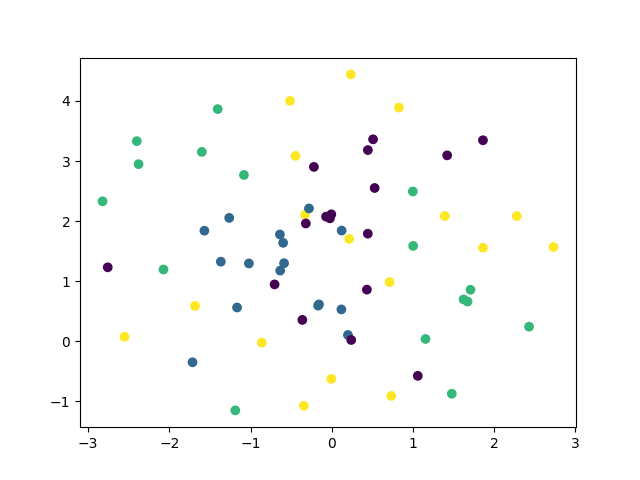}
    \caption{t-sne for hazelnut before contrastive classifier}
    \label{fig:hazelnut-a}
  \end{subfigure}
  \hfill
  \begin{subfigure}{0.49\linewidth}
    \includegraphics[width=\linewidth]{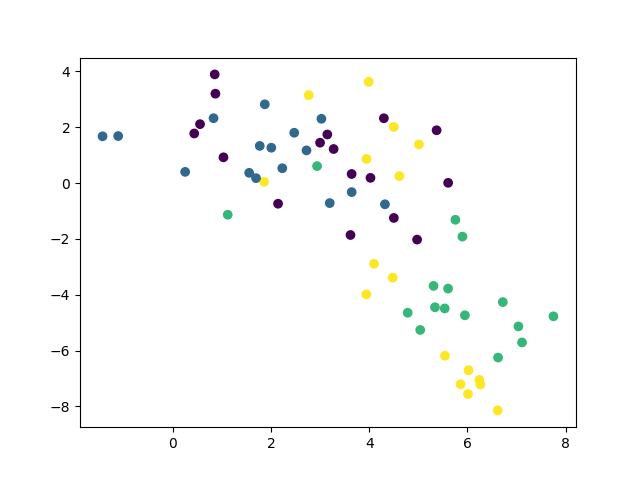}
    \caption{t-sne for hazelnut after contrastive classifier}
    \label{fig:hazelnut-b}
  \end{subfigure}
  \caption{t-sne for hazelnut classification}
  \label{fig:hazelnut}
\end{figure}

\begin{figure}[tb]
  \centering
  \begin{subfigure}{0.49\linewidth}
    \includegraphics[width=\linewidth]{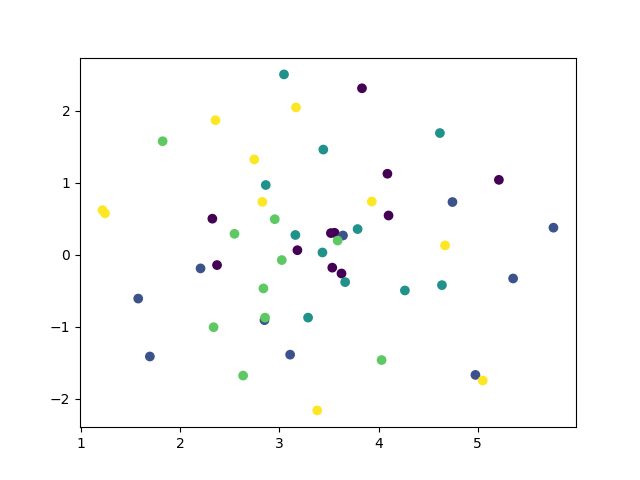}
    \caption{t-sne for grid before contrastive classifier}
    \label{fig:grid-a}
  \end{subfigure}
  \hfill
  \begin{subfigure}{0.49\linewidth}
    \includegraphics[width=\linewidth]{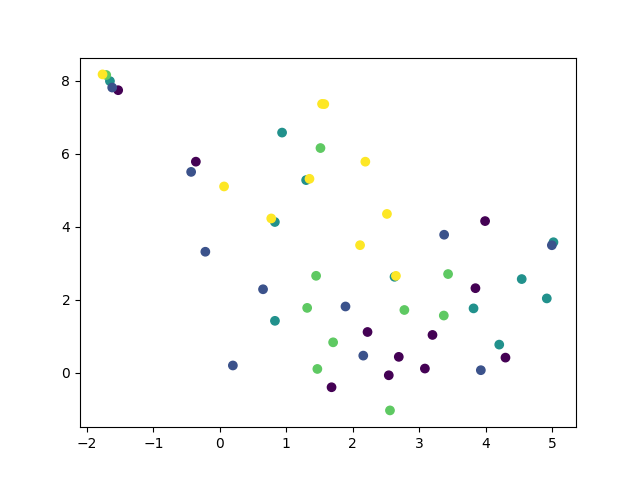}
    \caption{t-sne for grid after contrastive classifier}
    \label{fig:grid-b}
  \end{subfigure}
  \caption{t-sne for grid classification}
  \label{fig:grid}
\end{figure}

\begin{figure}[tb]
  \centering
  \begin{subfigure}{0.49\linewidth}
    \includegraphics[width=\linewidth]{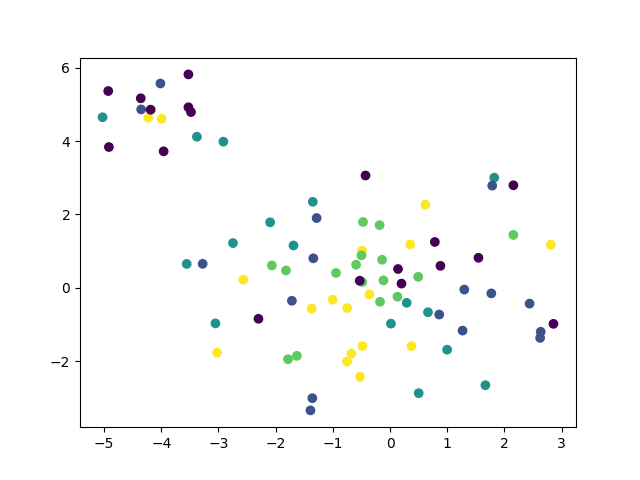}
    \caption{t-sne for carpet before contrastive classifier}
    \label{fig:carpet-a}
  \end{subfigure}
  \hfill
  \begin{subfigure}{0.49\linewidth}
    \includegraphics[width=\linewidth]{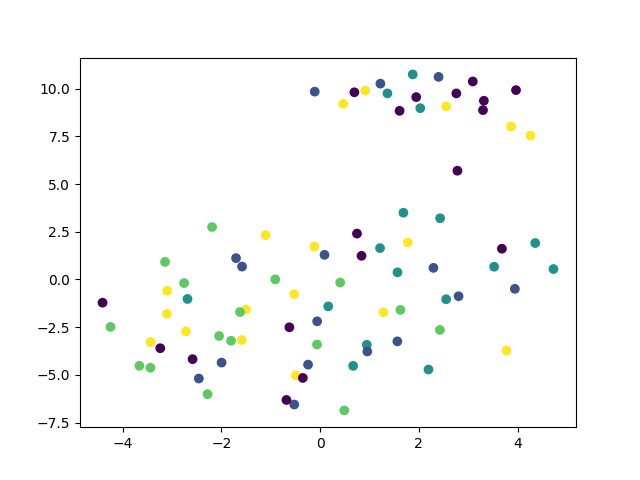}
    \caption{t-sne for carpet after contrastive classifier}
    \label{fig:carpet-b}
  \end{subfigure}
  \caption{t-sne for carpet classification}
  \label{fig:carpet}
\end{figure}

\begin{figure}[tb]
  \centering
  \begin{subfigure}{0.49\linewidth}
    \includegraphics[width=\linewidth]{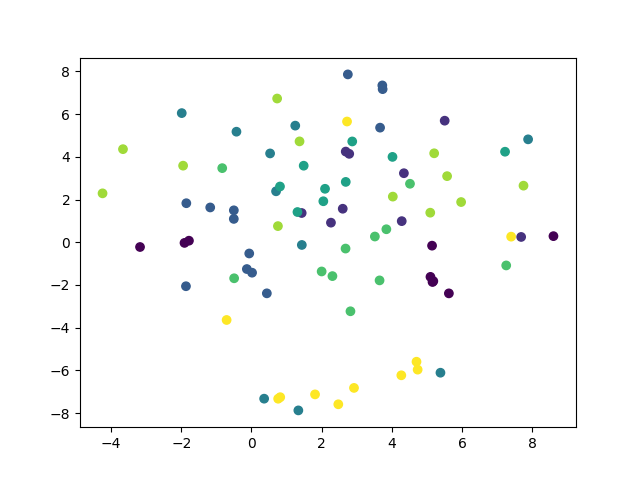}
    \caption{t-sne for cable before contrastive classifier}
    \label{fig:cable-a}
  \end{subfigure}
  \hfill
  \begin{subfigure}{0.49\linewidth}
    \includegraphics[width=\linewidth]{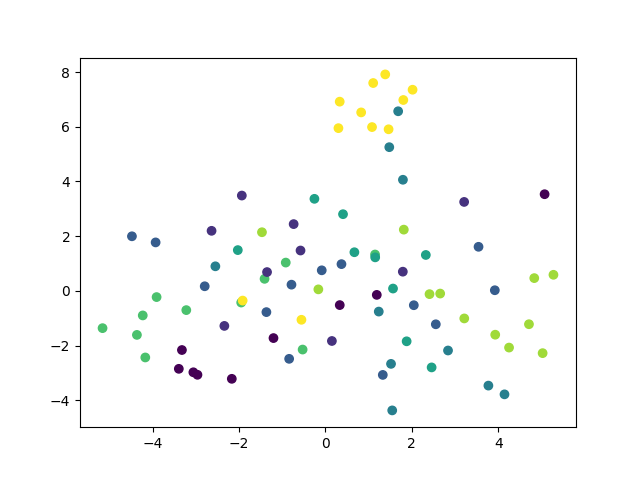}
    \caption{t-sne for cable after contrastive classifier}
    \label{fig:cable-b}
  \end{subfigure}
  \caption{t-sne for cable classification}
  \label{fig:cable}
\end{figure}

\begin{figure}[tb]
  \centering
  \begin{subfigure}{0.49\linewidth}
    \includegraphics[width=\linewidth]{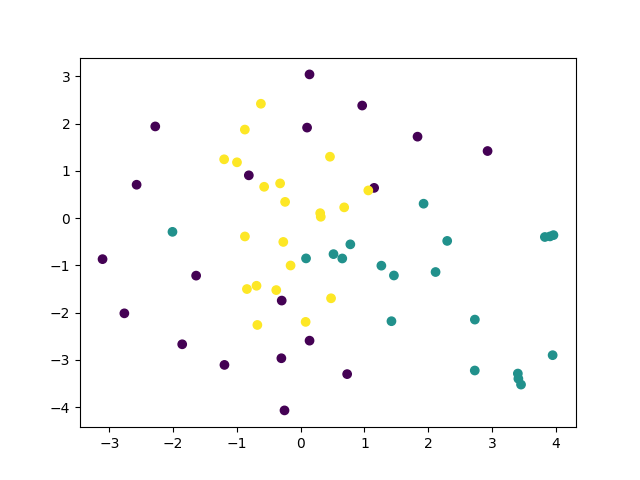}
    \caption{t-sne for bottle before contrastive classifier}
    \label{fig:bottle-a}
  \end{subfigure}
  \hfill
  \begin{subfigure}{0.49\linewidth}
    \includegraphics[width=\linewidth]{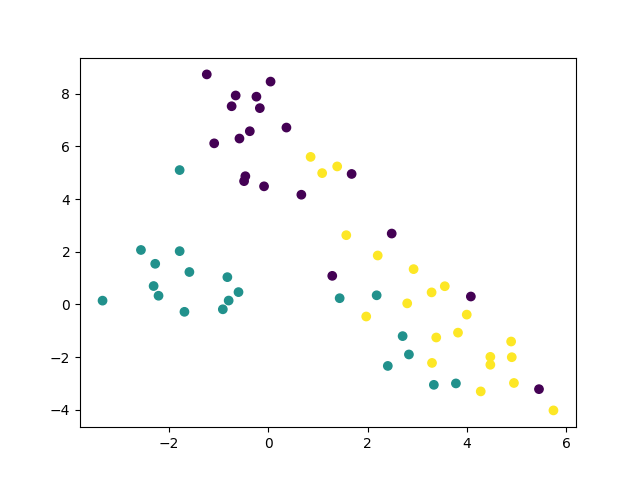}
    \caption{t-sne for bottle after contrastive classifier}
    \label{fig:bottle-b}
  \end{subfigure}
  \caption{t-sne for bottle classification}
  \label{fig:bottle}
\end{figure}

\begin{figure}[tb]
  \centering
  \begin{subfigure}{0.49\linewidth}
    \includegraphics[width=\linewidth]{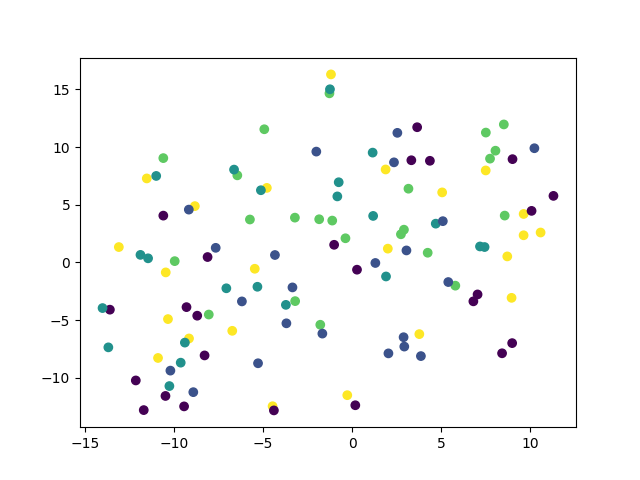}
    \caption{t-sne for screw before contrastive classifier}
    \label{fig:screw-a}
  \end{subfigure}
  \hfill
  \begin{subfigure}{0.49\linewidth}
    \includegraphics[width=\linewidth]{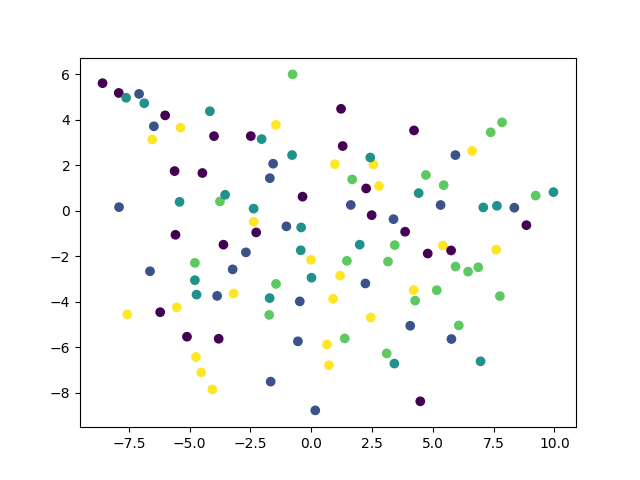}
    \caption{t-sne for screw after contrastive classifier}
    \label{fig:screw-b}
  \end{subfigure}
  \caption{t-sne for screw classification}
  \label{fig:screw}
\end{figure}

\begin{figure}[tb]
  \centering
  \begin{subfigure}{0.49\linewidth}
    \includegraphics[width=\linewidth]{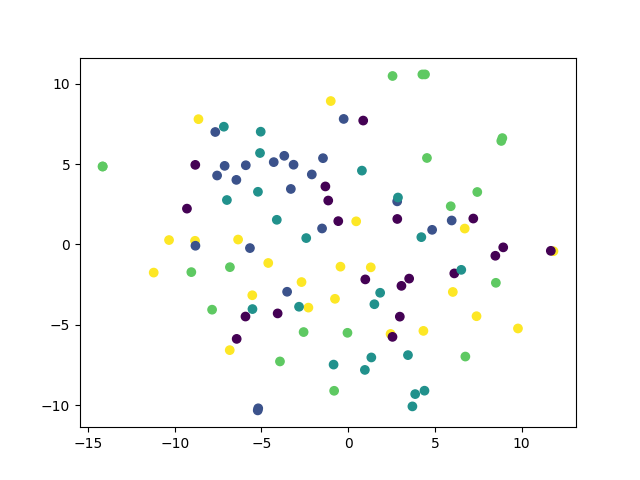}
    \caption{t-sne for capsule before contrastive classifier}
    \label{fig:capsule-a}
  \end{subfigure}
  \hfill
  \begin{subfigure}{0.49\linewidth}
    \includegraphics[width=\linewidth]{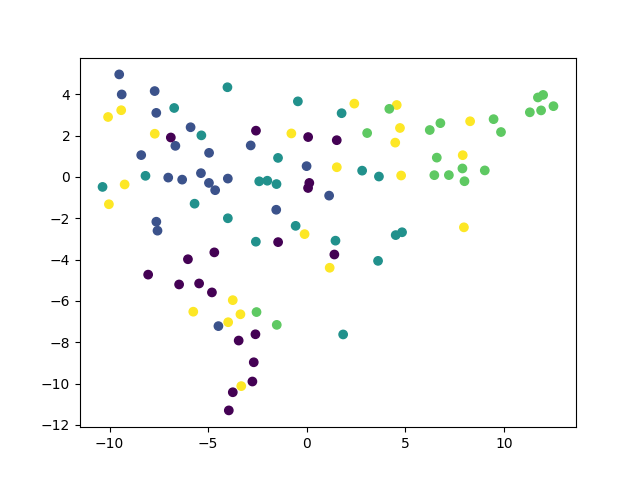}
    \caption{t-sne for capsule after contrastive classifier}
    \label{fig:capsule-b}
  \end{subfigure}
  \caption{t-sne for capsule classification}
  \label{fig:capsule}
\end{figure}

We will show the tSNE results of the remaining 12 types of items. Only hazelnut, metal nut, bottle and capsule shows significant performance. As shown from Figure \ref{fig:zipper} to Figure \ref{fig:capsule}

\begin{algorithm}[H]

\caption{Training Method for Vanilla Baseline}
\label{algorithm_vanilla}
\KwData{Input residual features $\mathcal{M}$, shot number $S$, classes number $C$}

\BlankLine
\tcp{Sort features by category}
$\mathcal{M}_{sorted} \leftarrow \{\}$\;
\For{$i=1$ \KwTo $C$}{
    \For{$j=1$ \KwTo $S$}{
    Extract $j$th feature of category $i$ from $\mathcal{M}$ and append to $\mathcal{M}_{sorted}$\;
    }
}
$\mathcal{M} \leftarrow \mathcal{M}_{sorted}$\;
\BlankLine

$\mathcal{S} \leftarrow \{\}$\;
\For{$i=1$ \KwTo $C$}{
    Extract One feature $\mathcal{F}$ of category $i$ from $\mathcal{M}$\;
    $ \mathcal{S} \leftarrow \mathcal{S} \cup \{ \mathcal{F}\}$\;
    $ \mathcal{M} \leftarrow \mathcal{M} \setminus \mathcal{F}$\;
    }

\tcp{Iterate over all remainning features in $\mathcal{M}$}
\For{each feature $\mathcal{F}$ in $\mathcal{M}$}{
    \tcp{Select current feature as the query sample and delete it from the bank}
    $\mathcal{Q} \leftarrow \mathcal{F}$\;
    $\mathcal{M} \leftarrow \mathcal{M} \setminus \mathcal{F}$\;
    \BlankLine
    \tcp{Now, $\mathcal{Q}$ and $\mathcal{S}$ can be used for training or evaluation}
}
\end{algorithm}
\begin{table}[h]
  \centering
  \caption{Complete Results of Vanilla Baseline}
  \label{tab:vanilla}
  \begin{tabular}{cccccc}
    \toprule
    category & two-shot(\%) & three-shot(\%) & four-shot(\%) & five-shot(\%) & one-shot(\%)\\
    \midrule
    bottle &  61.41 {\tiny $\pm 4.96$} & 52.59 {\tiny $\pm 12.18$} & 48.63 {\tiny $\pm 11.30$} & 40.00 {\tiny $\pm 4.75$} & 32.00 {\tiny $\pm 0.74$} \\
    cable & 15.53 {\tiny $\pm 3.27$} & 18.53 {\tiny $\pm 3.83$} & 19.00 {\tiny $\pm 4.50$} &  18.46 {\tiny $\pm 3.22$} & 13.57 {\tiny $\pm 4.26$}\\
    capsule &  22.02 {\tiny $\pm 2.41$} & 20.43 {\tiny $\pm 1.39$} & 20.22 {\tiny $\pm 2.43$} &  21.18 {\tiny $\pm 0.52$} & 17.50 {\tiny $\pm 0.80$}\\
    carpet &  24.05 {\tiny $\pm 5.30$} &  20.00 {\tiny $\pm 2.93$} &  17.97 {\tiny $\pm 2.20$} & 18.13 {\tiny $\pm 3.60$} & 21.43 {\tiny $\pm 3.37$}\\
    grid & 18.72 {\tiny $\pm 3.16$} & 21.43 {\tiny $\pm 3.76$} & 21.62 {\tiny $\pm 7.88$} &  18.13 {\tiny $\pm 4.64$} & 19.62 {\tiny $\pm 5.16$}\\
    hazelnut & 33.23 {\tiny $\pm 3.34$} & 32.07 {\tiny $\pm 3.97$} &  28.89 {\tiny $\pm 4.46$} & 33.60 {\tiny $\pm 3.85$} & 26.97 {\tiny $\pm 1.97$}\\
    leather & 21.22 {\tiny $\pm 6.77$} & 19.48 {\tiny $\pm 2.60$} &  23.33 {\tiny $\pm 2.28$} &  20.60 {\tiny $\pm 2.87$} & 23.45 {\tiny $\pm 2.09$}\\
    metal nut & 47.76 {\tiny $\pm 4.52$} & 51.11 {\tiny $\pm 2.56$} &  48.57 {\tiny $\pm 5.92$} & 51.51 {\tiny $\pm 3.43$} & 30.00 {\tiny $\pm 4.27$}\\
    pill &  20.63 {\tiny $\pm 4.18$} & 20.83 {\tiny $\pm 2.20$} & 19.65 {\tiny $\pm 3.56$} &  18.87 {\tiny $\pm 3.89$} & 6.12 {\tiny $\pm 0.82$}\\
    screw & 19.27 {\tiny $\pm 2.34$} &  20.00 {\tiny $\pm 2.19$} & 21.01 {\tiny $\pm 0.45$} &  18.94 {\tiny $\pm 2.31$} & 18.77 {\tiny $\pm 1.00$}\\
    tile &  35.41 {\tiny $\pm 6.37$} &  26.09 {\tiny $\pm 5.52$} &  25.31 {\tiny $\pm 2.04$} & 27.12 {\tiny $\pm 5.75$} & 24.56 {\tiny $\pm 2.91$}\\
    transistor & 32.51 {\tiny $\pm 8.15$} & 36.43 {\tiny $\pm 11.22$} & 36.67 {\tiny $\pm 11.56$} &  32.00 {\tiny $\pm 10.37$} & 27.78 {\tiny $\pm 2.78$}\\
    wood & 14.40 {\tiny $\pm 4.98$} & 19.11 {\tiny $\pm 9.89$} & 10.50 {\tiny $\pm 3.26$} &  26.86 {\tiny $\pm 15.20$} & 14.91 {\tiny $\pm 1.99$}\\
    zipper & 15.48 {\tiny $\pm 2.11$} & 13.06 {\tiny $\pm 3.18$} &  13.19 {\tiny $\pm 2.05$} & 13.34 {\tiny $\pm 1.00$} & 14.11 {\tiny $\pm 0.98$}\\
    Average & 27.43 {\tiny $\pm 1.05$} &  26.51 {\tiny $\pm 1.55$} & 25.42 {\tiny $\pm 1.37$} &  25.62 {\tiny $\pm 2.11$} & 20.77 {\tiny $\pm 0.77$}\\
    \bottomrule
  \end{tabular}
\end{table}

\begin{table}[tb]
  \centering
  \caption{Complete Results of Vanilla Baseline without Pretraining}
  \label{tab:vanilla_without_pretraining}
  \begin{tabular}{ccccc}
    \toprule
    category & two-shot(\%) & three-shot(\%) & four-shot(\%) & five-shot(\%)\\
    \midrule
    bottle &  39.04 {\tiny $\pm 7.64$} & 39.26 {\tiny $\pm 6.06$} & 37.25 {\tiny $\pm 4.60$} & 41.67 {\tiny $\pm 8.72$} \\
    cable & 19.03 {\tiny $\pm 2.86$} & 18.24 {\tiny $\pm 7.96$} & 14.33 {\tiny $\pm 4.95$} &  15.38 {\tiny $\pm 4.90$}\\
    capsule & 23.63 {\tiny $\pm 2.21$} & 20.43 {\tiny $\pm 1.90$} & 22.92 {\tiny $\pm 5.60$} &  21.19 {\tiny $\pm 5.28$}\\
    carpet & 21.52 {\tiny $\pm 4.10$} &  19.19 {\tiny $\pm 4.82$} &  18.84 {\tiny $\pm 2.90$} & 20.94 {\tiny $\pm 3.24$}\\
    grid & 22.13 {\tiny $\pm 4.90$} & 22.86 {\tiny $\pm 5.73$} & 19.46 {\tiny $\pm 6.45$} &  16.88 {\tiny $\pm 2.79$}\\
    hazelnut & 32.76 {\tiny $\pm 6.71$} & 30.26 {\tiny $\pm 3.29$} &  31.85 {\tiny $\pm 4.79$} & 29.47 {\tiny $\pm 2.72$}\\
    leather & 20.16 {\tiny $\pm 2.70$} & 19.22 {\tiny $\pm 3.24$} &  21.11 {\tiny $\pm 3.01$} &  18.81 {\tiny $\pm 4.08$}\\
    metal nut & 42.82 {\tiny $\pm 9.79$} & 49.88 {\tiny $\pm 2.24$} &  50.39 {\tiny $\pm 1.93$} & 49.32 {\tiny $\pm 4.33$}\\
    pill &  15.12 {\tiny $\pm 5.58$} & 13.50 {\tiny $\pm 6.70$} & 12.74 {\tiny $\pm 5.79$} &  18.11 {\tiny $\pm 4.08$}\\
    screw & 18.35 {\tiny $\pm 1.59$} &  23.73 {\tiny $\pm 3.80$} & 19.80 {\tiny $\pm 3.82$} &  20.42 {\tiny $\pm 3.64$}\\
    tile &  33.24 {\tiny $\pm 3.65$} &  30.44 {\tiny $\pm 4.10$} &  27.23 {\tiny $\pm 6.48$} & 27.46 {\tiny $\pm 4.55$}\\
    transistor & 25.00 {\tiny $\pm 9.63$} & 43.57 {\tiny $\pm 11.68$} & 39.17 {\tiny $\pm 9.13$} &  30.00 {\tiny $\pm 5.00$}\\
    wood & 22.80 {\tiny $\pm 7.01$} & 29.78 {\tiny $\pm 9.37$} & 29.00 {\tiny $\pm 6.27$} &  26.86 {\tiny $\pm 6.88$}\\
    zipper & 17.33 {\tiny $\pm 2.64$} & 12.65 {\tiny $\pm 1.55$} &  15.42 {\tiny $\pm 4.37$} & 13.10 {\tiny $\pm 2.23$}\\
    Average & 25.61 {\tiny $\pm 0.74$} &  28.12 {\tiny $\pm 1.40$} & 25.71 {\tiny $\pm 0.81$} &  25.63 {\tiny $\pm 1.17$}\\
    \bottomrule
  \end{tabular}
\end{table}

\begin{table}[tb]
  \centering
  \caption{Contrastive-Classifier Results in MvTec AD Dataset by new data generation method}
  \label{tab:new
  _data_generation}
  \begin{tabular}{ccccc}
  \toprule
  category & two-shot(\%) & three-shot(\%) & four-shot(\%) & five-shot(\%) \\
  \midrule
  bottle &  67.37 {\tiny $\pm 2.66$} &  74.07 {\tiny $\pm 3.93$} & 71.77 {\tiny $\pm 6.29$} & 76.67 {\tiny $\pm 6.81$}  \\
  cable & 30.00 {\tiny $\pm 5.22$} &  32.63 {\tiny $\pm 6.17$} & 34.33 {\tiny $\pm 6.08$} &  39.91 {\tiny $\pm 5.06$} \\
  capsule &  40.00 {\tiny $\pm 1.97$} &  39.79 {\tiny $\pm 3.73$} & 41.12 {\tiny $\pm 2.82$} &  45.24 {\tiny $\pm 4.69$}\\
  carpet &  23.54 {\tiny $\pm 2.12$} &  22.97 {\tiny $\pm 6.19$} &  24.93 {\tiny $\pm 6.27$} & 19.06 {\tiny $\pm 2.57$}\\
  grid & 27.66 {\tiny $\pm 1.51$} & 22.86 {\tiny $\pm 3.61$} & 22.16 {\tiny $\pm 4.84$} &  21.88 {\tiny $\pm 6.63$}\\
  hazelnut &  50.00 {\tiny $\pm 10.26$} &  41.38 {\tiny $\pm 2.44$} &  49.26 {\tiny $\pm 7.92$} & 57.20 {\tiny $\pm 5.40$}\\
  leather &  26.34 {\tiny $\pm 2.81$} &  31.43 {\tiny $\pm 4.34$} &  29.72 {\tiny $\pm 3.20$} &  31.34 {\tiny $\pm 6.76$}\\
  metal nut &  47.77 {\tiny $\pm 1.34$} & 49.38 {\tiny $\pm 3.38$} &  52.21 {\tiny $\pm 1.42$} & 53.70 {\tiny $\pm 3.12$}\\
  pill &  30.55 {\tiny $\pm 4.74$} & 29.50 {\tiny $\pm 3.04$} & 29.03 {\tiny $\pm 1.70$} &  28.68 {\tiny $\pm 0.52$}\\
  screw & 24.22 {\tiny $\pm 2.95$} &  21.35 {\tiny $\pm 1.26$} & 20.20 {\tiny $\pm 2.02$} &  20.00 {\tiny $\pm 3.87$}\\
  tile &  42.70 {\tiny $\pm 8.58$} &  57.10 {\tiny $\pm 6.20$} &  45.00 {\tiny $\pm 12.37$} & 42.03 {\tiny $\pm 15.41$}\\
  transistor & 35.00 {\tiny $\pm 6.40$} & 63.57 {\tiny $\pm 8.15$} & 55.83 {\tiny $\pm 10.87$} &  53.00 {\tiny $\pm 12.55$}\\
  wood & 31.60 {\tiny $\pm 5.90$} & 36.45 {\tiny $\pm 10.49$} & 39.50 {\tiny $\pm 8.73$} &  34.86 {\tiny $\pm 15.96$}\\
  zipper &  28.00 {\tiny $\pm 7.05$} & 20.82 {\tiny $\pm 2.66$} &  23.52 {\tiny $\pm 6.44$} & 20.24 {\tiny $\pm 3.67$}\\
  Average &  36.05 {\tiny $\pm 1.30$} &  38.81 {\tiny $\pm 0.71$} & 38.47 {\tiny $\pm 1.96$} &  38.80 {\tiny $\pm 2.10$}\\
  \bottomrule
\end{tabular}
\end{table}
\begin{table}[tb]
  \centering
  \caption{Complete Results of Direct Classification}
  \label{tab:direct}
  \begin{tabular}{cccccc}
    \toprule
    category & two-shot(\%) & three-shot(\%) & four-shot(\%) & five-shot(\%) & one-shot(\%)\\
    \midrule
    bottle & 26.32 & 29.63 & 39.22 & 33.33 & 33.33\\
    cable & 10.53 & 14.71 & 8.33 & 19.23 & 15.48\\
    capsule & 22.22 & 23.40 & 33.71 &  29.76 & 23.08\\
    carpet & 20.25 & 18.92 & 21.74 & 23.44 & 20.24\\
    grid & 21.28 & 21.43 & 21.62 & 21.88 & 19.23\\
    hazelnut & 33.87 & 31.03 & 38.89 & 38.00 & 25.76\\
    leather & 14.63 & 22.08 & 16.67 & 22.39 & 21.84\\
    metal nut & 28.24 & 37.04 & 28.57 & 32.88 & 17.98\\
    pill & 21.26 & 15.83 & 19.47 & 16.04 & 14.18\\
    screw & 16.51 & 18.27 & 16.16 & 20.21 & 16.67\\
    tile & 17.57 & 23.19 & 31.25 & 37.29 & 21.51\\
    transistor & 6.25 & 21.43 & 20.83 & 25.00 & 25.00\\
    wood & 32.00 & 24.44 & 15.00 & 25.71 & 18.18\\
    zipper & 18.10 & 16.33 & 15.38 & 21.43 & 16.07\\
    Average & 20.65 & 22.70 & 23.35 &  26.19 & 20.61 \\
    \bottomrule
  \end{tabular}
\end{table}
\begin{table}[tb]
  \centering
  \caption{The Complete Results of Contrastive Classifier without Residual Representation}
  \label{tab:without_residual}
  \begin{tabular}{ccccccc}
  \toprule
  category & two-shot(\%) & three-shot(\%) & four-shot(\%) & five-shot(\%) & one-shot(\%)\\
  \midrule
  bottle &  50.88 &  70.37 & 68.63 & 66.67 & 43.33 \\
  cable & 25.00 &  17.65 & 11.67 &  13.46 & 11.90\\
  capsule &  24.24 &  30.85 & 26.97 &  33.33 & 20.18\\
  carpet &  20.25 &  22.97 &  23.19 & 23.44 & 21.43\\
  grid & 34.04 & 23.81 & 21.62 &  18.75 & 19.23\\
  hazelnut &  50.00 &  36.21 &  37.04 & 30.00 & 31.82\\
  leather &  29.27 &  31.17 &  19.44 &  32.84 & 14.94 \\
  metal nut &  50.59 & 48.15 &  50.65 & 27.40 & 47.19 \\
  pill &  18.90 & 18.33 & 20.35 &  19.81 & 20.15 \\
  screw & 22.02 &  22.12 & 21.21 &  21.28 & 20.18 \\
  tile &  35.14 &  33.33 &  35.94 & 30.51 & 27.85 \\
  transistor & 56.25 & 64.29 & 45.83 &  35.00 & 38.89 \\
  wood & 44.00 & 46.67 & 40.00 &  22.86 & 27.27 \\
  zipper &  23.81 & 26.53 &  15.38 & 17.86 & 17.86 \\
  Average &  34.60 &  35.18 & 31.28 &  28.09 & 25.87 \\
  \bottomrule
\end{tabular}
\end{table}
\begin{table}[tb]
  \centering
  \caption{Results in MvTec 3D AD Dataset including normal samples}
  \label{tab:mvtec3d_good}
  \begin{tabular}{ccccccc}
    \toprule
    category & two-shot(\%) & three-shot(\%) & four-shot(\%) & five-shot(\%) & one-shot(\%)\\
    \midrule
    bagel & 24.60 {\tiny $\pm 4.22$} & 26.10 {\tiny $\pm 3.60$} & 22.67 {\tiny $\pm 2.17$} & 28.47{\tiny $\pm 1.53$} & 24.95 {\tiny $\pm 2.17$} \\
    cable gland & 33.47 {\tiny $\pm 3.98$} & 23.66 {\tiny $\pm 1.32$} & 31.81 {\tiny $\pm 1.61$} & 34.46 {\tiny $\pm 1.37$} & 20.00 {\tiny $\pm 2.13$}\\
    carrot & 24.22 {\tiny $\pm 4.47$} & 29.08 {\tiny $\pm 7.39$} & 29.19 {\tiny $\pm 3.53$} & 29.46 {\tiny $\pm 3.80$} & 24.71 {\tiny $\pm 2.33$}\\
    cookie & 29.75 {\tiny $\pm 2.48$} & 29.65 {\tiny $\pm 4.38$} & 33.15 {\tiny $\pm 3.40$} & 33.02 {\tiny $\pm 3.13$} & 20.32 {\tiny $\pm 2.67$}\\
    dowel & 26.83 {\tiny $\pm 4.22$} & 25.22 {\tiny $\pm 2.04$} &  33.64 {\tiny $\pm 4.90$} &  35.81 {\tiny $\pm 4.93$} & 24.48 {\tiny $\pm 3.78$}\\
    foam & 34.89 {\tiny $\pm 2.31$} &  38.12 {\tiny $\pm 2.95$} &  34.50 {\tiny $\pm 3.60$} & 30.67 {\tiny $\pm 6.11$} & 21.68 {\tiny $\pm 1.41$}\\
    peach & 29.18 {\tiny $\pm 1.89$} & 27.86 {\tiny $\pm 3.34$} &  27.32 {\tiny $\pm 2.72$} & 27.85 {\tiny $\pm 2.02$} & 22.83 {\tiny $\pm 1.47$} \\
    potato & 27.50 {\tiny $\pm 3.23$} & 35.35 {\tiny $\pm 2.47$} &  34.68 {\tiny $\pm 4.09$} &  36.18 {\tiny $\pm 2.30$} & 22.20 {\tiny $\pm 4.56$} \\
    rope & 49.25 {\tiny $\pm 11.08$} & 48.76 {\tiny $\pm 7.27$} & 61.88 {\tiny $\pm 10.04$} &  67.41 {\tiny $\pm 4.06$} & 31.34 {\tiny $\pm 1.17$} \\
    tire & 34.90 {\tiny $\pm 3.58$} & 35.26 {\tiny $\pm 4.27$} &  35.00 {\tiny $\pm 4.44$} &  - & 23.74 {\tiny $\pm 4.41$} \\
    Average & 31.46 {\tiny $\pm 1.56$} & 31.91 {\tiny $\pm 1.91$} & 34.38 {\tiny $\pm 1.94$} & 35.92 {\tiny $\pm 1.21$} & 23.63 {\tiny $\pm 1.05$} \\
    \bottomrule
  \end{tabular}
\end{table}
\begin{table}[tb]
  \centering
  \caption{Contrastive-Classifier Results in MvTec AD Dataset With MAML Training Tricks.}
  \label{tab:maml}
  \begin{tabular}{ccccccc}
  \toprule
  category & two-shot(\%) & three-shot(\%) & four-shot(\%) & five-shot(\%) & one-shot(\%)\\
  \midrule
  bottle &  68.07 {\tiny $\pm 3.80$} &  75.93 {\tiny $\pm 1.31$} & 83.14 {\tiny $\pm 4.72$} & 77.50 {\tiny $\pm 4.52$} & 55.67 {\tiny $\pm 3.03$} \\
  cable & 32.90 {\tiny $\pm 3.84$} &  32.94 {\tiny $\pm 4.37$} & 40.00 {\tiny $\pm 4.08$} &  38.85 {\tiny $\pm 9.06$} & 18.57 {\tiny $\pm 4.34$}\\
  capsule &  38.78 {\tiny $\pm 2.33$} &  36.38 {\tiny $\pm 2.05$} & 38.88 {\tiny $\pm 2.58$} &  44.05 {\tiny $\pm 4.29$} & 25.38 {\tiny $\pm 0.86$}\\
  carpet &  25.57 {\tiny $\pm 1.06$} &  26.49 {\tiny $\pm 2.63$} &  25.51 {\tiny $\pm 4.17$} & 24.06 {\tiny $\pm 2.37$}  & 21.43 {\tiny $\pm 3.15$}\\
  grid & 21.28 {\tiny $\pm 0.00$} & 23.33 {\tiny $\pm 3.91$} & 21.08 {\tiny $\pm 2.96$} &  20.00 {\tiny $\pm 1.71$} & 16.92 {\tiny $\pm 3.44$}\\
  hazelnut &  37.74 {\tiny $\pm 7.70$} &  28.97 {\tiny $\pm 2.56$} &  40.00 {\tiny $\pm 6.75$} & 48.80 {\tiny $\pm 11.01$} & 28.48 {\tiny $\pm 3.92$}\\
  leather &  23.66 {\tiny $\pm 1.85$} &  32.99 {\tiny $\pm 1.97$} &  35.28 {\tiny $\pm 2.33$} &  29.71 {\tiny $\pm 6.76$} & 19.54 {\tiny $\pm 2.30$} \\
  metal nut &  48.00 {\tiny $\pm 2.26$} & 49.38 {\tiny $\pm 1.51$} &  52.99 {\tiny $\pm 3.10$} & 56.98 {\tiny $\pm 2.67$} & 51.46 {\tiny $\pm 2.16$} \\
  pill &  32.60 {\tiny $\pm 4.12$} & 30.50 {\tiny $\pm 2.74$} & 28.67 {\tiny $\pm 1.34$} &  27.55 {\tiny $\pm 1.03$} & 23.73 {\tiny $\pm 3.09$} \\
  screw & 24.77 {\tiny $\pm 3.84$} &  23.46 {\tiny $\pm 3.01$} & 19.39 {\tiny $\pm 4.60$} &  20.07 {\tiny $\pm 4.26$} & 20.70 {\tiny $\pm 4.33$} \\
  tile &  47.30 {\tiny $\pm 4.48$} &  53.62 {\tiny $\pm 4.91$} &  54.38 {\tiny $\pm 4.04$} & 56.95 {\tiny $\pm 4.25$} & 25.32 {\tiny $\pm 5.14$} \\
  transistor & 56.25 {\tiny $\pm 3.83$} & 66.43 {\tiny $\pm 4.07$} & 60.83 {\tiny $\pm 6.97$} &  66.00 {\tiny $\pm 7.42$} & 35.00 {\tiny $\pm 6.69$} \\
  wood & 42.40 {\tiny $\pm 10.99$} & 33.78 {\tiny $\pm 8.23$} & 49.50 {\tiny $\pm 7.16$} &  37.71 {\tiny $\pm 9.35$} & 23.64 {\tiny $\pm 7.82$} \\
  zipper &  27.24 {\tiny $\pm 2.48$} & 28.98 {\tiny $\pm 2.94$} &  40.22 {\tiny $\pm 4.51$} & 23.10 {\tiny $\pm 4.80$} & 14.46 {\tiny $\pm 2.63$} \\
  Average &  37.61 {\tiny $\pm 0.93$} &  38.80 {\tiny $\pm 0.76$} & 42.13 {\tiny $\pm 0.81$} &  41.16 {\tiny $\pm 2.19$} & 27.17 {\tiny $\pm 1.15$} \\
  \bottomrule
\end{tabular}
\end{table}


\end{document}